\documentclass[runningheads]{llncs}

\usepackage[mobile]{eccv}

\usepackage{eccvabbrv}

\usepackage{graphicx}
\usepackage{booktabs}
\usepackage{multirow}
\usepackage[accsupp]{axessibility}  

\usepackage{comment}
\usepackage{bbm}

\usepackage{mathtools}
\usepackage{adjustbox}

\definecolor{skyblue}{RGB}{0, 191, 255}
\providecommand\given{}

\newcommand\SetSymbol[1][]{
    \nonscript\:#1\vert
    \allowbreak
    \nonscript\:
    \mathopen{}}

\DeclarePairedDelimiterX\Set[1]\{\}{
    \renewcommand\given{\SetSymbol[\delimsize]}
    #1
}

\usepackage{amsmath}
\DeclareMathOperator*{\argmaxA}{arg\,max} 

\usepackage{hyperref}

\usepackage{orcidlink}

\hypersetup{
    colorlinks=true,
    filecolor=cyan,      
    urlcolor=magenta,
}

\begin{document}

\title{Click-Gaussian: Interactive Segmentation\\to Any 3D Gaussians} 

\titlerunning{Click-Gaussian: Interactive Segmentation to Any 3D Gaussians}

\author{
Seokhun Choi$^1$\textsuperscript{*} \and
Hyeonseop Song$^1$\textsuperscript{*} \and \\
Jaechul Kim$^1$ \and
Taehyeong Kim$^2$\textsuperscript{\dag} \and
Hoseok Do$^1$\textsuperscript{\dag}
}

\authorrunning{S.~Choi \& H.~Song et al.}

%
\institute{
AI Lab, CTO Division, LG Electronics, Republic of Korea \\
\email{\{seokhun.choi, hyeonseop.song, jaechul1220.kim, hoseok.do\}@lge.com} \and
Dept. of Biosystems Engineering, Seoul National University, Republic of Korea \\
\email{taehyeong.kim@snu.ac.kr}  \\
Project page: \href{http://seokhunchoi.github.io/Click-Gaussian}{https://seokhunchoi.github.io/Click-Gaussian} \\
}
\maketitle
\let\thefootnote\relax\footnotetext{\textsuperscript{*}Equal contribution.}
\let\thefootnote\relax\footnotetext{\textsuperscript{\dag}Co-corresponding author.}

\begin{abstract}
Interactive segmentation of 3D Gaussians opens a great opportunity for real-time manipulation of 3D scenes thanks to the real-time rendering capability of 3D Gaussian Splatting. However, the current methods suffer from time-consuming post-processing to deal with noisy segmentation output. Also, they struggle to provide detailed segmentation, which is important for fine-grained manipulation of 3D scenes. In this study, we propose Click-Gaussian, which learns distinguishable feature fields of two-level granularity, facilitating segmentation without time-consuming post-processing. We delve into challenges stemming from inconsistently learned feature fields resulting from 2D segmentation obtained independently from a 3D scene. 3D segmentation accuracy deteriorates when 2D segmentation results across the views, primary cues for 3D segmentation, are in conflict. To overcome these issues, we propose Global Feature-guided Learning (GFL). GFL constructs the clusters of global feature candidates from noisy 2D segments across the views, which smooths out noises when training the features of 3D Gaussians. Our method runs in 10~ms per click, 15 to 130 times as fast as the previous methods, while also significantly improving segmentation accuracy.
\end{abstract}

\section{Introduction}

\label{sec:intro}

Recent progress in neural rendering technologies, such as Neural Radiance Fields (NeRF)\cite{mildenhall2021nerf}, along with novel 3D scene representation methods like 3D Gaussian Splatting (3DGS)~\cite{kerbl3Dgaussians}, have significantly impacted the field of photorealistic image synthesis within complex 3D environments.
These innovations extend into practical applications, enabling advancements in diverse domains such as virtual and augmented reality~\cite{jiang2024vr-gs, VRNeRF}, digital content creation~\cite{ren2023dreamgaussian4d, ling2023align, zielonka2023drivable}, and real-time rendering~\cite{chen2022mobilenerf} for interactive systems that demand both high fidelity and efficiency.
For these applications, accurately and efficiently segmenting objects within scenes is important~\cite{jiang2024vr-gs}, and presents ongoing challenges, particularly in distinguishing elements within diverse 3D environments.
\begin{figure*}[t]
\begin{center}
\centerline{\includegraphics[width=0.96\textwidth]
{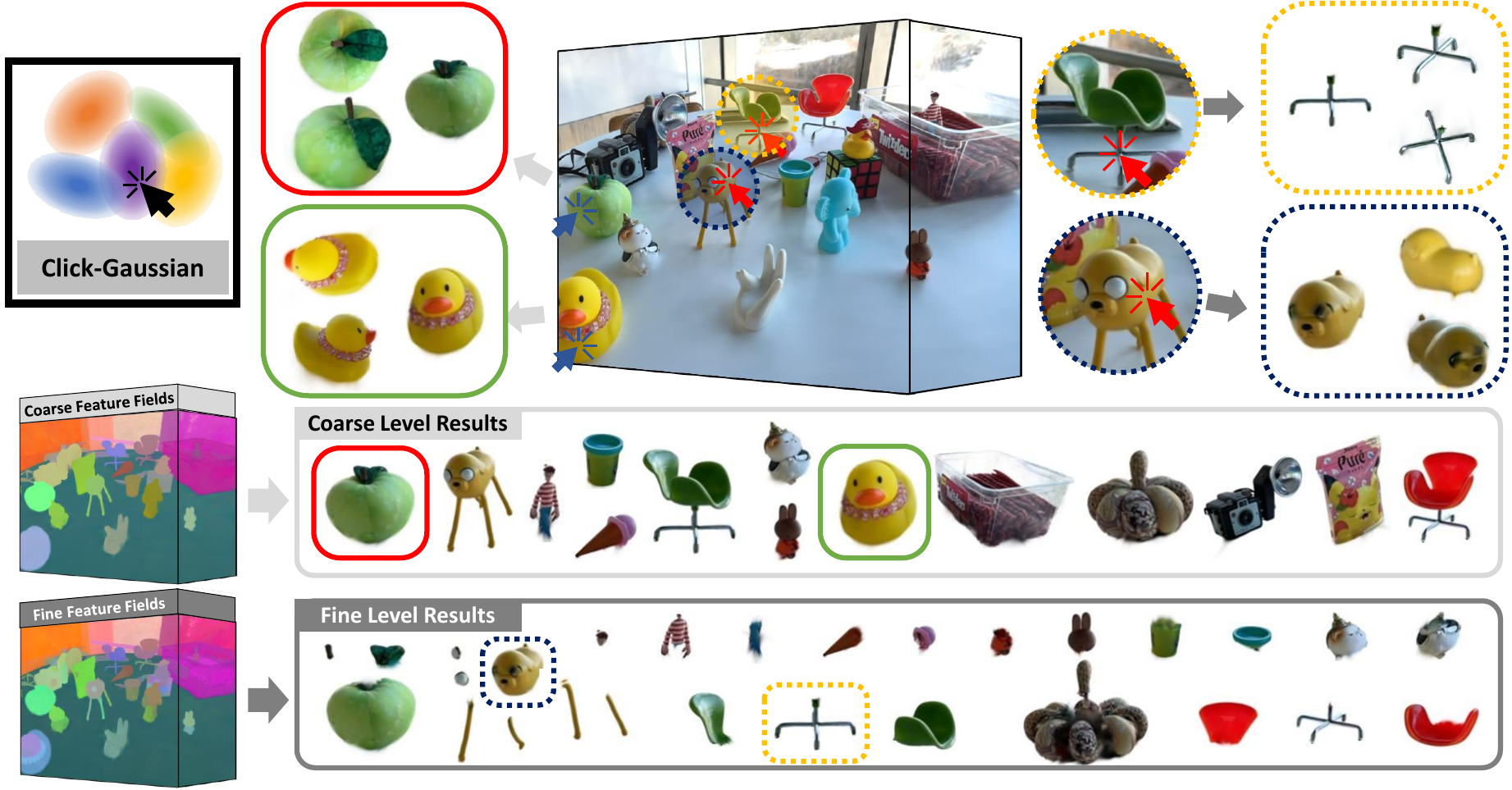}}
\vskip 0.1in
\vspace{-10px}
\caption{
We present Click-Gaussian, a swift and precise method for interactive segmentation of 3D Gaussians using two-level granularity feature fields derived from 2D segmentation masks.
Once trained, it enables users to select and segment desired objects at coarse and fine levels with a single click, completing the process within 10 ms.
}
\vspace{-27px}
\label{fig:teaser}
\end{center}
\end{figure*}

Recently, various segmentation methods~\cite{zhou2023feature, cen2024segment3dgaussians, ye2023gaussian} based on 3DGS have been proposed, leveraging advantages of 3DGS such as enhanced rendering efficiency and superior reconstruction quality.
For instance, some methods~\cite{zhou2023feature, cen2024segment3dgaussians} learn feature fields of 3D Gaussians that are aligned to the semantic representations from a foundational model like Segment Anything Model (SAM)~\cite{ICCV2023_SAM}.
This approach enables explicit segmentation of 3D scenes via 3D feature fields, which is crucial for supporting real-time applications and ensuring precise manipulation of intricate environments across diverse tasks.
However, these methods face challenges in learning distinguishable feature fields in a scene, necessitating extensive post-processing to achieve clear segmentation.
This reliance on time-consuming post-processing significantly impedes the efficiency benefits of 3DGS, creating a bottleneck for applications requiring rapid and direct manipulation of 3D scenes.
An alternative approach~\cite{ye2023gaussian} addresses 3D segmentation by utilizing object tracking mechanisms~\cite{cheng2023tracking} to pre-assign SAM-based segment identities.
However, this method's efficacy is contingent on successful object tracking, potentially excluding untracked objects from the segmentation process.
This limitation suggests the potential benefit of developing more robust segmentation techniques capable of comprehensively handling diverse objects within complex scenes.

These considerations motivate the exploration of 3D segmentation methods that provide distinguishable feature fields without extensive post-processing.
Progress in these techniques could significantly improve real-time interaction with 3D scenes, thereby enabling more intuitive and responsive experiences in 3D object manipulation tasks.
Such advancements have the potential to not only enhance 3D scene editing capabilities but also broaden the practical applications of 3D scene representation across diverse fields.

In this study, we propose Click-Gaussian, depicted in Fig.~\ref{fig:teaser}, as a practical and efficient method for interactive segmentation of 3D Gaussians pre-trained on real-world scenes.
By elevating 2D segmentation masks extracted from the SAM into enriched 3D Gaussian's augmented features with two-level granularity, \ie, coarse and fine levels, our approach facilitates fine-grained segmentation.
The two-level granularity enables Click-Gaussian to capture scene elements at different scales in 3D environments, enhancing the precision and details of segmentation outcomes.
This is achieved through the incorporation of a granularity prior for Gaussian's features, coupled with the employment of a contrastive feature learning method based on the SAM's 2D segmentation masks.
Additionally, a significant hurdle in this process is the inconsistency of 2D masks across different views, which impedes the training of consistent and distinguishable semantic features.
To address this issue, we introduce Global Feature-guided Learning (GFL), a novel strategy that systematically aggregates global feature candidates across the training views to coherently inform the development of 3D feature fields.
The GFL technique enhances the robustness and reliability of feature learning, mitigating the impact of inherent ambiguities present in individual 2D segmentation masks.
We demonstrate the effectiveness of Click-Gaussian through comprehensive experiments on complex real-world scenes, evaluating both segmentation accuracy and computational efficiency.
Our results indicate that this approach offers a promising solution for precise and rapid 3D scene manipulation, potentially facilitating applications across various domains.

In summary, our key contributions are as follows:
\begin{itemize}
\item We propose Click-Gaussian, which enables interactive segmentation of any 3D Gaussians by utilizing two-level feature fields derived from 2D segmentation masks using contrastive learning methods and a granularity prior.
\item To tackle the issue of 2D mask inconsistency across training views, we propose GFL, a novel approach that gathers global feature candidates from an entire scene to consistently guide Gaussian's feature learning.
\item Through extensive experiments on complicated real-world scenes, we confirm the effectiveness of our approach, demonstrating its suitability for interactive segmentation by significantly enhancing accuracy and processing time.
\end{itemize}

\section{Related Work}

\subsubsection{3D Gaussian Representations.}
3D Gaussian Splatting~\cite{kerbl3Dgaussians} has emerged as a promising method for real-time scene rendering, offering superior visual quality. 
This has inspired research~\cite{luiten2023dynamic, yang2023deformable, yang2023real, Cotton_2024_WACV} into dynamic scene reconstruction, leveraging its fast rendering capabilities through the design of deformation fields~\cite{yang2023deformable, yang2023real, Cotton_2024_WACV}.
Moreover, the research has expanded into 3D~\cite{tang2023dreamgaussian, yi2023gaussiandreamer, chen2023text} and 4D~\cite{ren2023dreamgaussian4d, ling2023align} content generation by incorporating diffusion models~\cite{rombach2022high, liu2023zero}.
These studies demonstrate the efficient rendering and high visual fidelity of 3D Gaussian representations in various applications.
Our study further extends these capabilities by focusing on the segmentation of 3D Gaussians, while maintaining their inherent advantages.

\subsubsection{Feature Distillation for 3D Segmentation.}
Recent approaches to 3D segmentation can be broadly categorized into two main strategies: feature distillation and mask-lifting techniques.
Feature distillation approaches~\cite{kobayashi2022distilledfeaturefields, tschernezki22neuralyang2023real, isrfgoel2023, kerr2023lerf, chen2023interactive, zhou2023feature} aim to transfer high-dimensional features from 2D vision foundation models~\cite{radford2021learning, caron2021emerging} into 3D representations.
For instance, DFFs~\cite{kobayashi2022distilledfeaturefields}, N3F~\cite{tschernezki22neuralyang2023real}, and ISRF~\cite{isrfgoel2023} utilize DINO~\cite{caron2021emerging}, while LeRF~\cite{kerr2023lerf} employs CLIP~\cite{radford2021learning}.
However, these foundational models, not specifically designed for segmentation tasks, make such approaches struggle to achieve fine-grained segmentation.
More recent studies, like those by Chen~\textit{et al.}\cite{chen2023interactive} and Feature3DGS\cite{zhou2023feature}, distill SAM's encoder features into 3D and use the SAM's decoder to interpret 2D rendered feature maps for segmentation.
However, the computational demands of SAM's decoder limit real-time interactive 3D segmentation.
In contrast, our approach achieves finer segmentation in real-time by lifting SAM-generated 2D masks to 3D space.

\subsubsection{2D Mask-lifting for 3D Segmentation.}
In addition to feature distillation approaches, recent studies have explored lifting 2D segmentation masks into 3D space~\cite{cen2024segment, ren-cvpr2022-nvos, mirzaei2023spin, cen2024segment3dgaussians, ye2023gaussian, ying2023omniseg3d, kim2024garfield}.
For instance, SA3D~\cite{cen2024segment} and NVOS~\cite{ren-cvpr2022-nvos} utilize user prompts (\eg, points or scribbles) to derive segmentation masks for a target object in reference views, subsequently training a neural field with these masks for object segmentation.
MVseg~\cite{mirzaei2023spin} uses a video segmenter~\cite{caron2021emerging} to get multi-view masks.
While the aforementioned approaches focus on single-object segmentation, other studies~\cite{ying2023omniseg3d, cen2024segment3dgaussians, ye2023gaussian, kim2024garfield} have developed methods for segmenting multiple objects simultaneously.
OmniSeg3D~\cite{ying2023omniseg3d} trains a feature field using a hierarchical contrastive learning method with 2D segmentation masks, achieving fine-grained segmentation by adjusting cosine similarity thresholds.
GARField~\cite{kim2024garfield} addresses inconsistent SAM-generated masks across views by introducing a scale-conditioned feature field.
However, both use NeRF-based structures, which face computational challenges during rendering, limiting real-time performance.

In the context of 3DGS,  SAGA~\cite{cen2024segment3dgaussians} also employs a contrastive learning method with SAM-generated masks.
It projects SAM's features into a low-dimensional space via a trainable MLP, imitating these features to address inconsistency issues.
However, the distilled SAM's feature is detrimental to the segmentation method that uses feature cosine similarity at the inference stage, requiring extensive post-processing for accuracy.
Gau-Group~\cite{ye2023gaussian} takes a different approach, applying a zero-shot tracker~\cite{cheng2023tracking} to address mask inconsistencies under the assumption that training images form a video sequence.
This assumption, though, limits generalizability, and the method struggles with untracked SAM masks.
In contrast, our proposed Global Feature-guided Learning (GFL) method ensures view-consistent training signal by leveraging globally aggregated feature candidates throughout a scene without assuming sequential image inputs.

\section{Methods}
\begin{figure*}[t!]
\begin{center}
\centerline{\includegraphics[width=0.99\textwidth]
{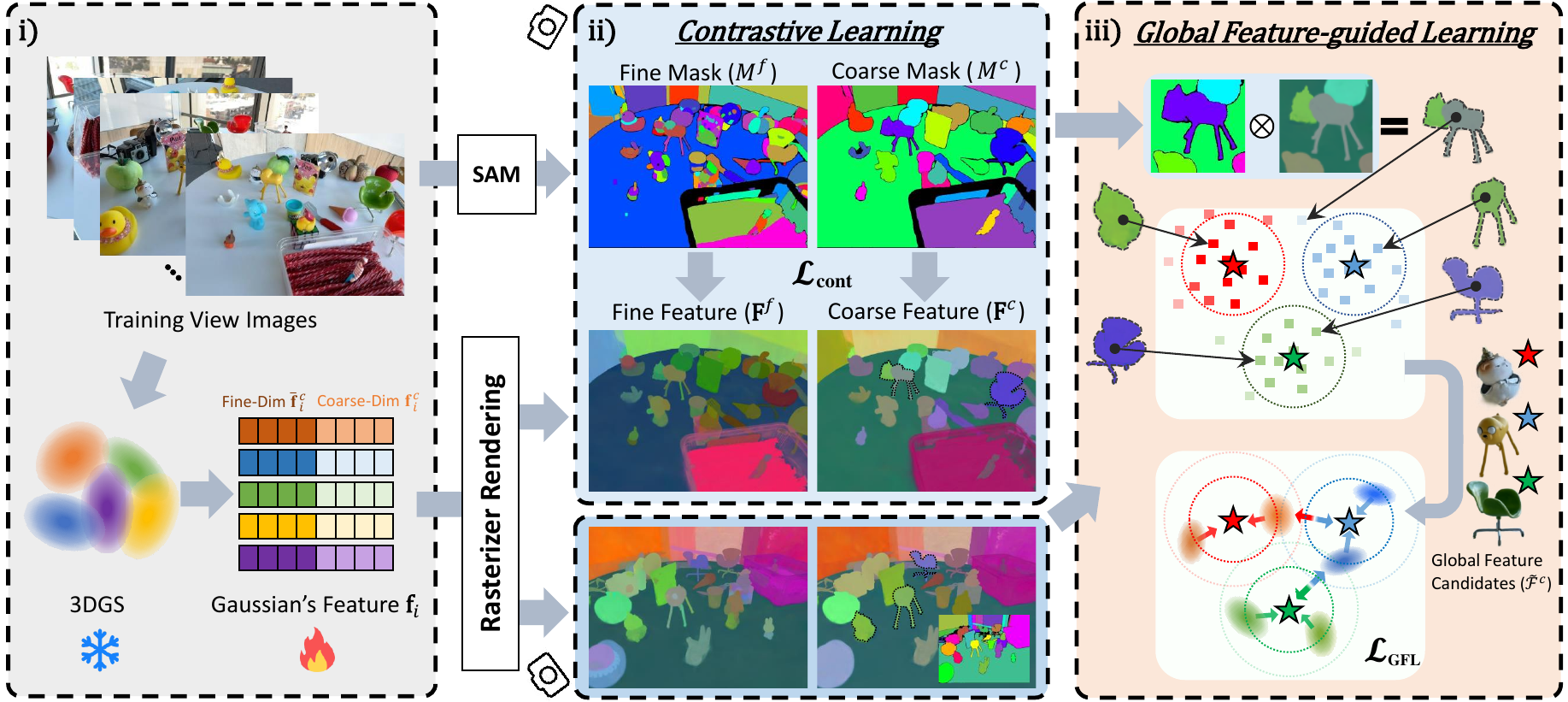}}
\vskip 0.1in
\vspace{-10px}
\caption{
Overview of the proposed method.
i) Our approach augments pre-trained 3D Gaussians with two-level granularity features $\mathbf{f}_i$. 
ii) These features are trained through contrastive learning, utilizing 2D rendered feature maps $\mathbf{F}$ and their corresponding SAM-generated masks $M$.
iii) To address inconsistencies in mask signals across views, we introduce a Global Feature-guided Learning approach.
For clarity, Global Feature-guided Learning at the fine level is omitted from the illustration.
}
\vspace{-25px}
\label{fig:method_overview}
\end{center}
\end{figure*}

We propose Click-Gaussian, a 3D segmentation method that augments pre-trained 3D Gaussians with effective and distinct 3D feature fields, enabling real-time segmentation capabilities for 3D Gaussian representations.
To achieve this, we initially utilize the \textit{automatic mask generation module} of SAM~\cite{ICCV2023_SAM} for all training views of a scene, then organize generated masks based on their segment areas to derive coarse and fine level masks for each image.
The information from these two-level masks is then incorporated into 3D Gaussians by splitting each Gaussian's feature space using a granularity prior, facilitating the representation of both levels of detail~(Sec.~\ref{sec:Click-Gaussian}).
We train these augmented features through contrastive learning, applied to 2D rendered feature maps in conjunction with the masks~(Sec.~\ref{sec:Contrastive Learning}).
To enhance the consistency of feature learning across different viewpoints, we propose \textbf{G}lobal \textbf{F}eature-guided \textbf{L}earning (GFL), which aggregates global feature candidates across the scene during training~(Sec.~\ref{sec:Global Feature-guided Learning}). 
Additionally, we employ several regularization methods in our training process to further stabilize and refine the training of Click-Gaussian's features~(Sec.~\ref{sec:Regularization}).
The comprehensive methodology is illustrated in Fig.~\ref{fig:method_overview}.

\subsection{Preliminary: 3D Gaussian Splatting} 
3D Gaussian Splatting (3DGS) represents a 3D scene with explicit 3D Gaussians and uses differentiable rasterizer~\cite{kerbl3Dgaussians} for rendering. 
Formally, given a training image set $\mathcal{I}=\{I^v\}_{v=1}^V$ with camera poses, it aims to learn a set of 3D Gaussians $G = \{g_{i}\}^{N}_{i=1}$, where $V$ is the number of training images, $N$ is the number of Gaussians, and $g_{i} = \{\mathbf{p}_i, \mathbf{s}_i, \mathbf{q}_i, o_i, \mathbf{c}_i\}$ is $i$-th Gaussian's trainable parameters. 
Here, $\mathbf{p}_i \in \mathbb{R}^3$ is each Gaussian's center position. 
A scaling factor $\mathbf{s}_i \in \mathbb{R}^3$ and a quaternion $\mathbf{q}_i \in \mathbb{R}^4$ are used to represent each Gaussian's 3D covariance. 
$o_i \in \mathbb{R}$ is an opacity value, and $\mathbf{c}_i$  is a color represented with spherical harmonics coefficients~\cite{yu_and_fridovichkeil2021plenoxels}.
After projecting 3D Gaussians onto 2D image space with a given camera pose, 3DGS uses the rasterizer to compute a color $\mathbf{C}$ on a pixel by performing $\alpha$-blending~\cite{kopanas2021point, kopanas2022neural} on $\mathcal{N}$ depth-ordered points overlapping the pixel:
\begin{equation}
\label{eq:pixel_color}
\mathbf{C} = \sum_{i \in \mathcal{N}} \mathbf{c}_i \alpha_i T_i,
\end{equation}
where $\alpha_i$ is calculated by evaluating the influences of each projected Gaussian using splatted 2D covariance~\cite{yifan2019differentiable}, opacity $o_i$, and pixel distance, and $T_i = \prod_{j=1}^{i-1} (1 - \alpha_j)$ is the transmittance.

\subsection{Feature Fields of Click-Gaussian}
\label{sec:Click-Gaussian}

Click-Gaussian operates by equipping each 3D Gaussian in a scene with additional features for segmentation.
Specifically, given a 3D Gaussian $g_{i}$, each Gaussian is augmented with a $D$-dimensional feature vector $\mathbf{f}_i \in \mathbb{R}^{D}$ for 3D segmentation, resulting in $\tilde{g_{i}}=g_{i} \cup \{\mathbf{f}_i\}$.
We split $\mathbf{f}_i$ into $\mathbf{f}_i^c \in \mathbb{R}^{D^c}$ and $\bar{\mathbf{f}}_i^c \in \mathbb{R}^{D-D^c}$, enabling Click-Gaussian to learn features well on both the coarse and fine level masks.
We use $\mathbf{f}_i^c$ as a coarse-level feature, and $\mathbf{f}_i^f = \mathbf{f}_i^c \oplus \bar{\mathbf{f}}_i^c$, not $\bar{\mathbf{f}}_i^c$, as a fine-level feature where $\oplus$ is a concatenate function.
This is motivated by the intrinsic dependency between two levels in the real world, called granularity prior (\eg, if two objects $A$ and $B$ are different at the coarse level, then each fine part $a \subset A$ and $b \subset B$ are naturally different), to make fine-level feature learning more effective.
For our experimental setup, we set $D^c=12$ and $D=24$ and freeze other parameters of Gaussians except features.
Using the rasterizer, we can compute two-level features $\mathbf{F}^l$ on a pixel, akin to the method outlined in Eq.~(\ref{eq:pixel_color}):
\begin{equation}
\label{eq:pixel_feature}
\mathbf{F}^l = \sum_{i \in \mathcal{N}} \mathbf{f}_i^l \alpha_i T_i,
\end{equation}
where $l=\{f, c\}$ is the granularity level.
The computation of two-level features for each pixel is conducted in a single forward pass.

\subsection{Contrastive Learning}
\label{sec:Contrastive Learning}
We use cosine similarity based contrastive learning to train distinctive features with a set of two-level masks.
To illustrate this concretely, consider a two-level mask $M^l \in \mathcal{M}$ for a training image $I \in \mathcal{I}$, where $l=\{f, c\}$ is the granularity level.
For pixels $p_1$ and $p_2$, if their mask values are the same, \ie, $M^l_{p_1}=M^l_{p_2}$, we aim to maximize the cosine similarity between their rendered features: 
\begin{equation}
\label{eq:loss_contrastive_positive}
\mathcal{L}_{\text{pos}}^{\text{cont}} = -\frac{1}{|P_1||P_2|} \sum_{l}^{\{f, c\}} \sum_{p_1}^{P_1}\sum_{p_2}^{P_2}\mathbbm{1}\left[M^l_{p_1}=M^l_{p_2}\right] \mathbf{S}^l(p_1, p_2),
\end{equation}
where $\mathbbm{1}$ is the indicator function, $P_1$ and $P_2$ are the set of sampled pixels, $| \cdot|$ is the number of elements in a set, and $\mathbf{S}^l(p_1, p_2)=\langle\mathbf{F}^l_{p_1}, \mathbf{F}^l_{p_2}\rangle$ is the cosine similarity between rendered features of two pixels. 
Conversely, for pixels with different mask values, \ie, $M^l_{p_1} \neq M^l_{p_2}$, we constrain their rendered features' cosine similarity to not exceed a specified margin, $\tau^l$:
\begin{equation}
\label{eq:loss_contrastive_negative}
\mathcal{L}_{\text{neg}}^{\text{cont}} = \frac{1}{|P_1||P_2|} \sum_{l}^{\{f, c\}} \sum_{p_1}^{P_1}\sum_{p_2}^{P_2}\mathbbm{1}\left[M^l_{p_1} \neq M^l_{p_2}\right] \mathbbm{1}\left[\mathbf{S}^l(p_1, p_2)>\tau^l\right] \mathbf{S}^l(p_1, p_2).
\end{equation}
Considering that two points may represent distinct parts at the fine level yet be classified as the same object at the coarse level, we apply stop gradient operations, \textit{sg}, to the coarse-level components during optimization for negative contrastive loss on fine-level features: $\mathbf{F}^f = \textit{sg}(\mathbf{F}^c) \oplus \bar{\mathbf{F}}^c$.
This method effectively focuses the training process on elements critical for discerning fine-level distinction.
We set the margins $\tau^f=0.75$ and $\tau^c=0.5$ for all experimental settings.
The total contrastive learning loss is defined as:
\begin{equation}
\label{eq:loss_contrastive}
\mathcal{L}_{\text{cont}} = \mathcal{L}_{\text{pos}}^{\text{cont}} + \lambda_{\text{neg}}^{\text{cont}}\mathcal{L}_{\text{neg}}^{\text{cont}},
\end{equation}
where $\lambda_{\text{neg}}^{\text{cont}}$ is a hyperparameter for balancing the two losses.

\subsection{Global Feature-guided Learning}
\label{sec:Global Feature-guided Learning}
Click-Gaussian's features, despite being trained through contrastive learning, face challenges due to inconsistencies in SAM-generated masks across training viewpoints.
This issue arises from the independent use of the masks in each view, potentially leading to unreliable training signals.
To address this, we propose Global Feature-guided Learning (GFL), a method that continuously acquires global feature candidates to provide non-conflicting and reliable supervision.

\subsubsection{Global Feature Candidates.} 
After a specified number of training iterations, we calculate the average features for each two-level mask across all training views.
This is accomplished by rendering 2D feature maps and applying average pooling to each mask for all training views.
Formally, for two-level masks for all viewpoints $\mathcal{M}^l=\{M^{l,v}\}_{v=1}^V$, where $l=\{f, c\}$ denotes granularity level, $V$ is the number of training viewpoints, and $M^{l,v}\ \in \mathbb{Z}^{H \times W}$ represents a mask for viewpoint $v$ at level $l$, the average features are calculated as follows:
\begin{equation}
\label{eq:average features}
\mathcal{F}^l = 
\Set*{\bar{F}_{s}^{l,v} \in \mathbb{R}^{D^l} \given \bar{F}_{s}^{l,v}=\frac{1}{|\mathcal{P}^{l,v}_{s}|} \sum_{p \in \mathcal{P}^{l,v}_{s}} \mathbf{F}_p^{l,v}, 1 \le s \le \max_v{M^{l,v}}}.
\end{equation}
Here, $\mathcal{P}^{l,v}_{s}=\Set*{p \given M_p^{l,v}=s}$ is a set of pixels with the same segment identiy (ID) in mask $M^{l,v}$, and $D^l$ is the feature dimension at level $l$.
This average pooling procedure is done rapidly without gradient calculation, thanks to the real-time rendering speed of 3DGS at inference time. 
We then obtain $C^l$ global feature candidates for each level, denoted as $\Tilde{\mathcal{F}}^l$, across a scene by applying the HDBSCAN~\cite{HDBSCAN_paper} clustering algorithm to each set $\mathcal{F}^l$.
These global feature candidates are periodically updated to obtain the latest global features.
Notably, as these global clusters are derived by grouping rendered features from noisy 2D segments across all views, they become the most representative features for the entire scene, effectively mitigating inconsistencies in the SAM-generated masks.

\subsubsection{Global Feature-guided Learning.}
Global feature candidates enable supervision of Click-Gaussian in a view-consistent manner.
For a Gaussian's feature $\mathbf{f}_i^l$, we guide the feature to belong to a specific global cluster and to be far from the others. 
This involves identifying $c^l_i$, the cluster ID where the $i$-th Gaussian's feature is most likely to belong at level $l$: $c^l_i=\argmaxA_{c} \Tilde{\mathbf{S}}^l(i, c)$, where $\Tilde{\mathbf{S}}^l(i, c) = \langle \mathbf{f}^l_i, \Tilde{\mathcal{F}}^l_c \rangle$ is the cosine similarity between the $i$-th Gaussian's feature and global cluster feature with ID $c$.
The GFL loss function for supervising the Gaussian's feature to belong to the most likely global cluster is defined as:
\begin{equation}
\label{eq:loss_GFL_pos}
\mathcal{L}_{\text{pos}}^{\text{GFL}} = -\frac{1}{N} \sum_{l}^{\{f, c\}} \sum_{i}^{N} \mathbbm{1}\left[\Tilde{\mathbf{S}}^l(i, c^l_i) > \tau^g \right] \Tilde{\mathbf{S}}^l(i, c^l_i).
\end{equation}
Here, $\tau^g$ is a threshold for determining whether to belong to the cluster, and we set $\tau^g=0.9$ across all our experiments.
Conversely, the GFL loss function guiding the Gaussian's feature away from other global clusters is defined as:
\begin{equation}
\label{eq:loss_GFL_neg}
\mathcal{L}_{\text{neg}}^{\text{GFL}} = \frac{1}{N} \sum_{l}^{\{f, c\}} \sum_{i}^{N} \frac{1}{C^l} \sum_{c\neq c^l_i}^{C^l} \mathbbm{1}\left[\Tilde{\mathbf{S}}^l(i, c) > \tau^l \right] \Tilde{\mathbf{S}}^l(i, c),
\end{equation}
where $\tau^l$ is described in Eq.~(\ref{eq:loss_contrastive_negative}).
The total GFL loss is thus formulated as:
\begin{equation}
\label{eq:loss_global_cluster}
\mathcal{L}_{\text{GFL}} = \mathcal{L}_{\text{pos}}^{\text{GFL}} + \mathcal{L}_{\text{neg}}^{\text{GFL}}.
\end{equation}
Applying the GFL loss directly to Gaussian's features using global clusters enhances their distinctiveness and noise robustness through reliable supervision, which is vital for accurate 3D segmentation as shown in Sec.~\ref{sec:ablation}.

\subsection{Regularization}
\label{sec:Regularization}
\subsubsection{Hypersphere Regularization.}
\label{sec:Hypersphere Regularization}
Features with excessively large norms underestimate the participation of other features in the rendering process in Eq.~(\ref{eq:pixel_feature}), impeding effective learning of all Gaussian's features.
To prevent any single Gaussian’s feature from dominating in the $\alpha$-blending process~\cite{kopanas2021point, kopanas2022neural} of the feature rendering, similar to~\cite{ying2023omniseg3d}, we constrain Gaussian's features to lie on the surface of the hypersphere:
\begin{equation}
\label{eq:3D-norm}
\mathcal{L}_{\text{3D-norm}} = \frac{1}{N} \sum_{i=1}^{N} \left({||\mathbf{f}_i^c||_2-1}\right)^2+\left({||\bar{\mathbf{f}}_i^c||_2-1}\right)^2.
\end{equation}
\subsubsection{Rendered Feature Regularization.}
Due to the hypersphere regularization in Eq.~(\ref{eq:3D-norm}), each Gaussian's feature of level $l$ lie on the surface of a hypersphere of radius $r^l$, with $r^c=1$ for coarse and $r^f=\sqrt 2$ for fine levels.
However, the norm of the rendered feature, $||\mathbf{F}^l_p||_2$, is less than $r^l$, as feature vectors $\mathbf{f}_i$ in different directions are integrated by Eq.~(\ref{eq:pixel_feature}).
This implies that Gaussian's features $\mathbf{f}_i$ contributing to $\mathbf{F}^l_p$ for a single pixel (\ie, the same object) can vary.
To ensure all contributing features $\mathbf{f}_i$ for rendering $\mathbf{F}^l_p$ are aligned in the same direction, we apply the following regularization on the rendered feature:
\begin{equation}
\label{eq:2D-norm}
\mathcal{L}_{\text{2D-norm}} = \frac{1}{HW} \sum_{l}^{\{f, c\}} \sum_{p}^{HW} \left({||\mathbf{F}^l_p||_2-r^l}\right)^2.
\end{equation}

\subsubsection{Spatial Consistency Regularization.}
Following the approach of \cite{ye2023gaussian}, we leverage 3D spatial information to ensure that proximate Gaussians exhibit similar features.
At the outset of training, we construct a KD-tree~\cite{kdtree_paper} using the 3D positions of Gaussians to facilitate efficient queries for spatially proximate neighbors.
Throughout the training process, we sample $N_s$ Gaussians and adjust their features to align with those of their $K$-nearest neighbors in 3D space:
\begin{equation}
\label{eq:3D-geo}
\mathcal{L}_{\text{spatial}} = -\frac{1}{N_s K} \sum_{i}^{N_s} \sum_{k}^{K} \langle  \mathbf{f}_i, \mathbf{f}_k \rangle,
\end{equation}
where $\langle  \cdot, \cdot \rangle$ is the cosine similarity operation.
For all experiments, we set $N_s=100,000$ and $K=5$.
Finally, our total objective for training Click-Gaussian is: 
\begin{equation}
\label{eq:loss_total}
\mathcal{L}_{\text{total}}=\mathcal{L}_{\text{cont}}+\lambda_1\mathcal{L}_{\text{GFL}}+\lambda_2\mathcal{L}_{\text{3D-norm}}+\lambda_3\mathcal{L}_{\text{2D-norm}}+\lambda_4\mathcal{L}_{\text{spatial}},
\end{equation}
where $\lambda_1$, $\lambda_2$, $\lambda_3$, and $\lambda_4$ are hyperparameters balancing the respective loss terms.

\section{Experiments}
\subsection{Experimental Settings}
We implemented Click-Gaussian using the 3DGS codebase~\cite{kerbl3Dgaussians}, adopting its default settings for pre-trained Gaussians.
The hyperparameters were set as follows: $\lambda_{\text{neg}}^{\text{cont}}=0.1$, $\lambda_1=10.0$, $\lambda_2=0.2$, $\lambda_3=0.2$, and $\lambda_4=0.5$.
We employed the Adam optimizer with a learning rate of $0.01$ for Gaussian's features.
For contrastive learning, we sampled 10k pixels for each training iteration using importance sampling based on mask pixel count.
In HDBSCAN, we set the epsilons for coarse and fine features clustering to $1\times 10^{-2}$ and $1\times 10^{-3}$, respectively, with the minimum cluster size proportional to the number of training views.
We trained Click-Gaussian for 3,000 iterations, incorporating Global Feature-Guided learning from the 2,000th iteration onward. 
The entire training process took approximately 13 minutes on an NVIDIA RTX A5000 GPU.

We used the official code’s \textit{automatic mask generation module} for SAM mask creation, which extracts masks without distinguishing levels, allowing us to get only the highest-confidence segments in an image.
The coarse and fine masks are then assigned per pixel by the largest and smallest segments, respectively (see supplementary materials for details).
To evaluate our approach's 3D segmentation performance using these masks, we employ two public real-world datasets: LERF-Mask dataset~\cite{ye2023gaussian} and SPIn-NeRF dataset~\cite{mirzaei2023spin}.
The LERF-Mask dataset comprises three scenes~\cite{kerr2023lerf} with manually annotated ground truth masks for large objects.
We further annotated several masks for fine objects within each scene using Make-Sense~\cite{make-sense} to evaluate fine-grained segmentation performance.
The SPIn-NeRF dataset offers multi-view masks for single objects in the widely used NeRF datasets~\cite{mildenhall2019local, mildenhall2021nerf, Tanks-and-Temples, yen2022nerfsupervision, yu2022plenoxels} that include both forward-facing~\cite{mildenhall2019local} and 360-degree inward-facing~\cite{mildenhall2021nerf, Tanks-and-Temples, yen2022nerfsupervision, yu2022plenoxels} setups.

\subsubsection{Segmentation Procedure.}
After training for each dataset, we computed global feature candidates for subsequent segmentation tasks.
We adopted a label propagation method~\cite{ye2023gaussian, cen2024segment, cen2024segment3dgaussians} for 3D multi-view segmentation task by using a ground truth 2D mask from a reference view to identify target object IDs and evaluated generated 2D masks.
Specifically, we calculated cosine similarities between the rendered 2D feature map and the pre-computed global clusters, applying a 0.9 threshold to determine cluster IDs of the target object.
Using the IDs, we generated 2D masks for test views and evaluated performance by calculating mIoU between rendered and ground truth masks.
This approach aligns with common scenarios where users segment a target object in the reference view to obtain results for other views.
For 3D Gaussian extraction task, we used cosine similarity between rendered features corresponding to user-provided point prompts and two-level global feature candidates to retrieve clicked objects efficiently.

\begin{table}[t!]
    \begin{center}
    \begin{adjustbox}{max width=0.99\textwidth}
        \begin{tabular}{c|ccc|ccc|ccc|ccc}
             \multirow{2}{*}{Model} &  \multicolumn{3}{c|}{Figurines} &  \multicolumn{3}{c|}{Ramen} &  \multicolumn{3}{c|}{Teatime} &  \multicolumn{3}{c}{Average}  \\
             & \scriptsize{Coarse} & \scriptsize{Fine} & \scriptsize{All} & \scriptsize{Coarse} & \scriptsize{Fine} & \scriptsize{All} & \scriptsize{Coarse} & \scriptsize{Fine} & \scriptsize{All} & \scriptsize{Coarse} & \scriptsize{Fine} & \scriptsize{All} \\             
            \hline
            Gau-Group~\cite{ye2023gaussian} & 69.7& 17.1& 33.8& 77.0& 75.1& 75.8& 71.7& 7.78& 43.3& 72.8& 31.5& 49.9 \\
            OmniSeg3D~\cite{ying2023omniseg3d} & 87.1& 28.0& 46.8& 77.3& 72.1& 74.2& 73.8& 46.9& 61.8& 79.4& 46.9& 60.9  \\
            Feature3DGS~\cite{zhou2023feature} & 70.4& 49.2& 55.9& 65.9& 73.0& 70.2& 60.6& 68.4& 64.0& 65.6& 63.5& 63.4  \\
            GARField~\cite{kim2024garfield} & 89.2& 59.4& 70.1& 75.7& \textbf{86.9}& 82.4& 77.8& 67.8& 73.2& 80.9& 71.4& 75.2  \\
            Ours & \textbf{93.2}& \textbf{75.9}& \textbf{81.4}& \textbf{90.9}& 86.7& \textbf{88.4}& \textbf{83.2}& \textbf{90.4}& \textbf{86.4} & \textbf{89.1}& \textbf{84.3}& \textbf{85.4}  \\ 
        \end{tabular}
    \end{adjustbox}
    \end{center}
    \vspace{-5px}
    \caption{
    Quantitative comparison with baselines on LERF-Mask Dataset.
    Segmentation performance was evaluated in mIoU with coarse and fine level masks for each of three scenes, and `All' reflects mean value across all objects and levels.
    `Average' shows mean performance across scenes.
    Our approach outperforms all prior works, especially in fine-level segmentation.
    }
    \label{table:lerf-mask-data-table}
\vspace{-15px}
\end{table}

\subsection{Comparisons}
\begin{figure*}[t!]
\begin{center}
\centerline{\includegraphics[width=0.95\textwidth]
{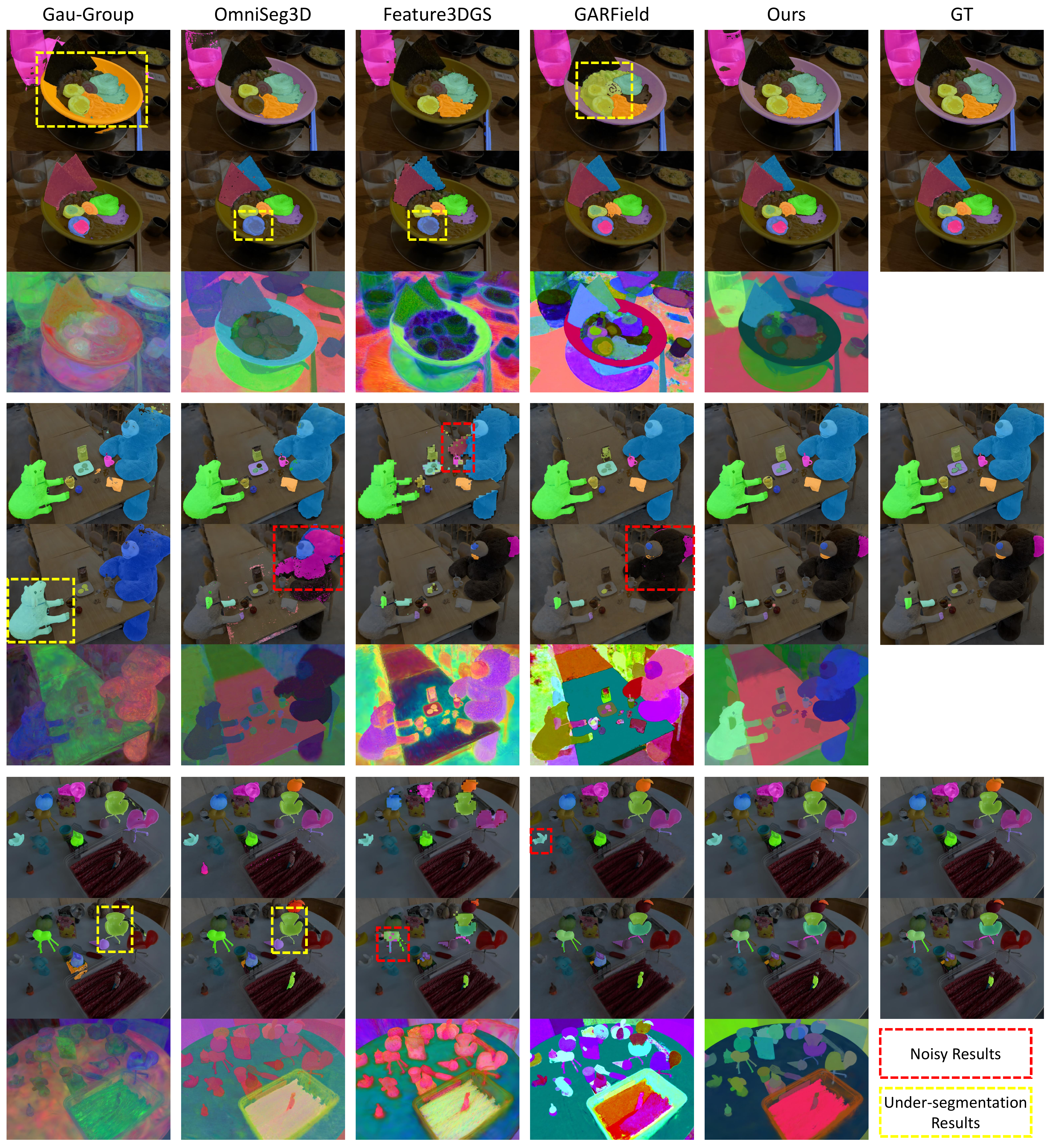}}
\vskip 0.1in
\vspace{-10px}
\caption{
Comparison with baselines on LERF-Mask Dataset. 
The results are displayed in three lines per scene (Teatime, Ramen, and Figurines in order). 
Each scene's first two rows show coarse and fine level segmentation results, respectively, and the third row shows the PCA visualizations of each model's finest-level feature field. 
Our approach demonstrates superior segmentation ability in both coarse and fine levels. 
Red and yellow boxes indicate noisy and under-segmentation results, respectively. 
}
\vspace{-33px}
\label{fig:lerf_data_comparison}
\end{center}
\end{figure*}

\subsubsection{Comparison on LERF-Mask Dataset.}
To demonstrate Click-Gaussian's segmentation superiority, we compared it with various baselines using the LERF-Mask dataset.
For Gau-Group, target object IDs in the reference view were identified using a classifier and a ground truth mask~\cite{ye2023gaussian}. 
OmniSeg3D's segmentation involved adjusting the cosine similarity threshold~\cite{ying2023omniseg3d} from 0 to 1 in 0.01 increments, finding the optimal threshold for each target object. 
Feature3DGS utilized a rendered 2D feature map and the SAM's decoder for segmentation~\cite{zhou2023feature}, selecting the best match for the target object. 
GARField applied a NeRF-based scale-conditioned field, selecting the best scale (0 to 1 in 0.05 steps) for each target object in the reference view~\cite{kim2024garfield}.

\begin{figure*}[t!]
\begin{center}
\centerline{\includegraphics[width=0.95\textwidth]
{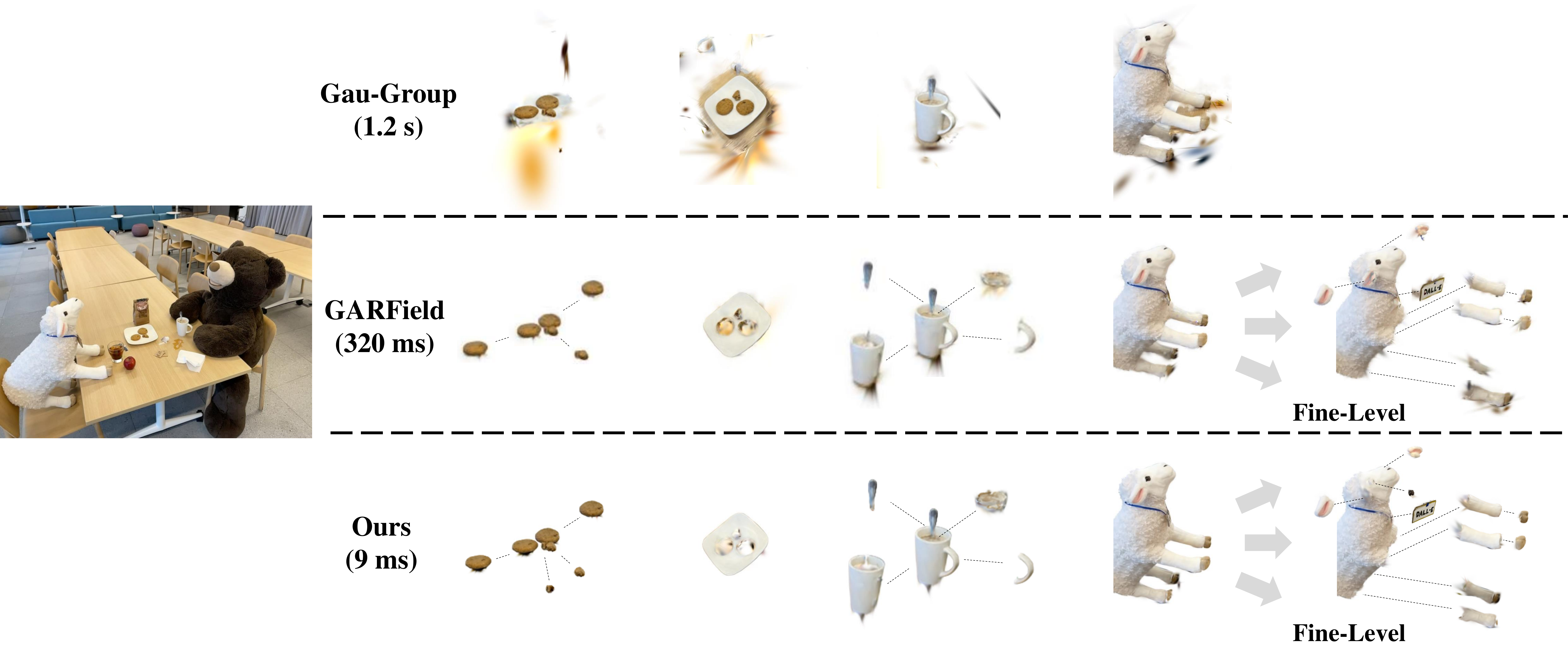}}
\vskip 0.1in
\vspace{-10px}
\caption{
Comparison with Gau-Group and GARField. 
Our approach performs more detailed and cleaner extractions of Gaussians, up to 130 times faster than other baselines.
}
\vspace{-25px}
\label{fig:teatime_gaugroup_garfield_comparison}
\end{center}
\end{figure*}

\begin{figure*}[t!]
\begin{center}
\centerline{\includegraphics[width=0.95\textwidth]
{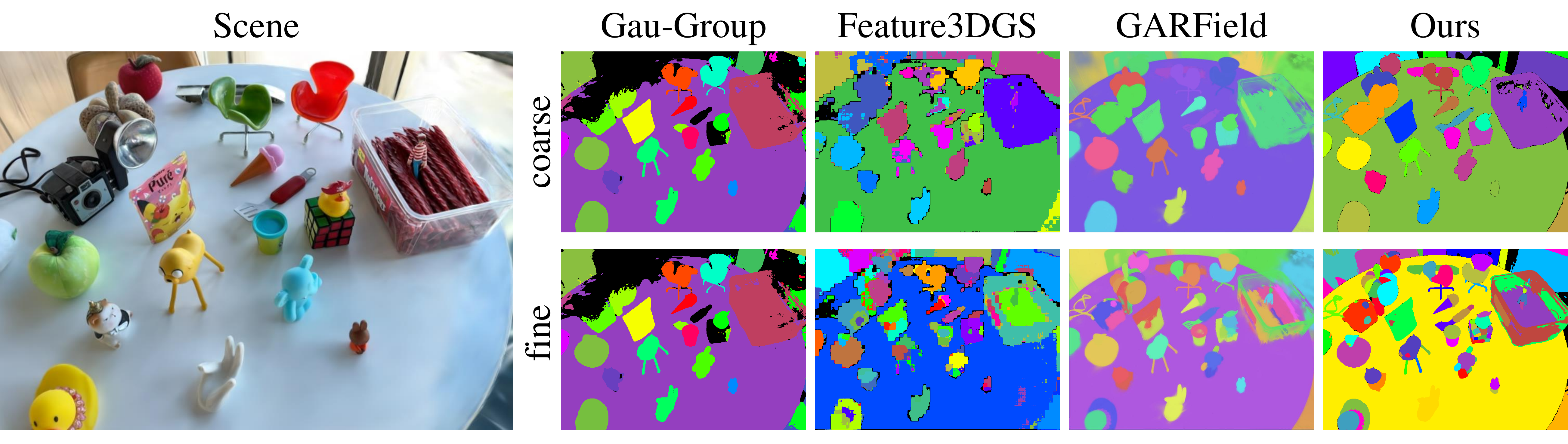}}
\vskip 0.1in
\vspace{-10px}
\caption{Comparison for automatic segmentation of everything on novel views.
Our method shows more exact and fine-grained results against baselines.
For detailed experimental procedures, please refer to the supplementary materials.
Gau-Group, unable to differentiate levels, presents identical coarse and fine segmentation results.
}
\vspace{-25px}
\label{fig:autogen}
\end{center}
\end{figure*}

As shown in Tab.~\ref{table:lerf-mask-data-table}, our method outperforms all baselines.
Gau-Group underperforms in fine-level segmentation, due to its tracking methodology limitations. 
OmniSeg3D shows improved performance than Gau-Group using a hierarchical field separated by cosine similarity, but lacks detailed segmentation.
Feature3DGS enhances performance via SAM's decoder but requires a longer processing time.
GARField shows comparable results with scale-conditioned feature fields, yet struggles with finer segmentation accuracy.
Fig.~\ref{fig:lerf_data_comparison} qualitatively demonstrates our method's superiority, together with PCA visualizations of the finest level feature fields of all models.

We also qualitatively compared 3D Gaussian extraction performance using user-provided point prompts.
As shown in Fig.~\ref{fig:teatime_gaugroup_garfield_comparison}, our method yields finer and cleaner extractions while operating up to 130 times faster than competing approaches.
Additionally, as depicted in Fig.~\ref{fig:autogen}, we compared performance in automatic segmentation of everything on a novel view, where Click-Gaussian shows more exact and fine-grained results than all baselines.

\subsubsection{Comparison on SPIn-NeRF Dataset.}
\begin{figure*}[t!]
\begin{center}
\centerline{\includegraphics[width=0.95\textwidth]
{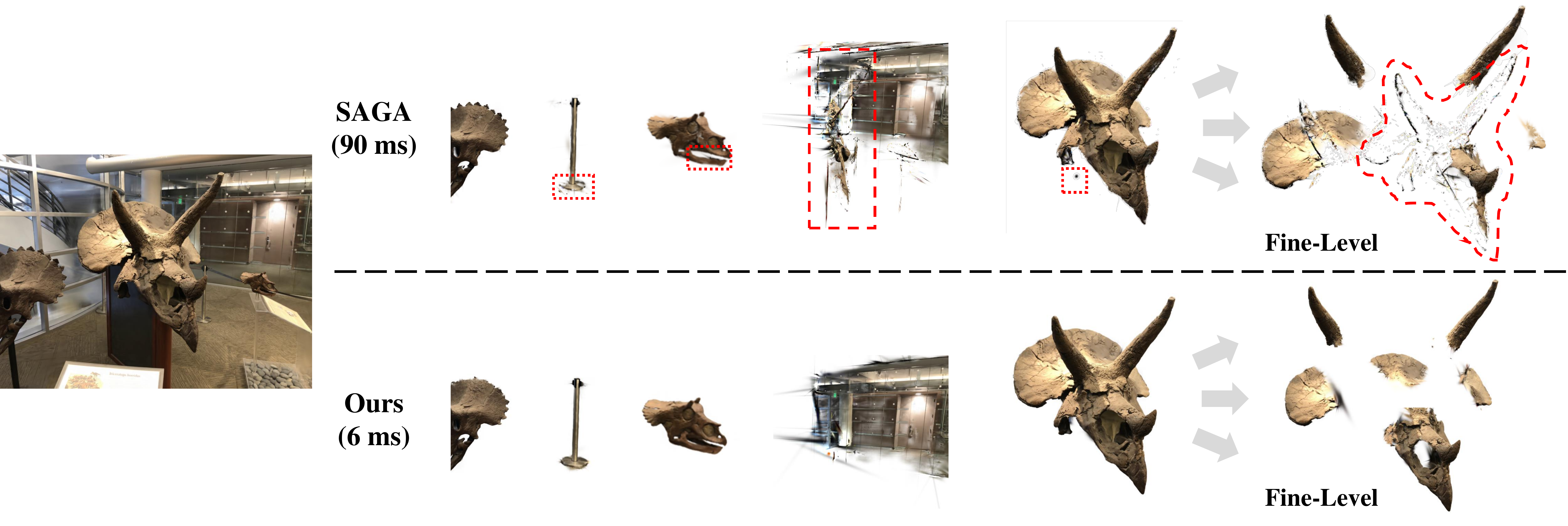}}
\vskip 0.1in
\vspace{-10px}
\caption{
Comparison with SAGA. Our method achieves more precise Gaussian extraction, highlighted by red dotted lines, and runs about 15 times faster than SAGA.
}
\vspace{-15px}
\label{fig:horn_saga_comparison}
\end{center}
\end{figure*}

We extend our analysis to include additional baselines on the SPIn-NeRF dataset, evaluating our method's performance across various views per scene.
The first two baseline models in Tab.~\ref{table:spinnerf-dataset-table}, MVSeg and SA3D, can segment only a single foreground object at a time.
Click-Gaussian, which efficiently segments any objects within a single training, even outperforms both methods.
It also surpasses SAGA, which utilizes low-dimensional SAM's feature fields.
In Fig.~\ref{fig:horn_saga_comparison}, we further experimented on 3D Gaussian extraction using user-provided point prompts.
Our method performed detailed and exact Gaussian extraction about 15 times faster than SAGA which uses time-consuming multiple post-processing steps~\cite{cen2024segment3dgaussians}.

\begin{table}[t!]
    \begin{center}
        \begin{tabular}{c|c|c|c|c}
             &  {~~MVSeg~\cite{mirzaei2023spin}~~} &  {~~SA3D~\cite{cen2024segment}~~} &  {~~SAGA~\cite{cen2024segment3dgaussians}~~} & {~Ours}\\
            \hline
            ~mIoU $(\%)$~ & 90.9& 92.4& 88.0& ~~\textbf{94.0}~~\\
        \end{tabular}
    \end{center}
    \caption{
     Quantitative comparison with baselines on SPIn-NeRF Dataset.
     We report average mIoU across ten real-world scenes.
     Our method outperforms not only MVSeg and SA3D, which segment only a single foreground object per training, but also SAGA.
    }
    \label{table:spinnerf-dataset-table}
\vspace{-15px}
\end{table}

\begin{table}[t!]
    \begin{center}
        \begin{tabular}{c|c|c|c|c|c|c}
              & {\scriptsize{w/o} $\mathcal{L}_{\text{2D-norm}}$} & {\scriptsize{w/o} $\mathcal{L}_{\text{3D-norm}}$} & {\scriptsize{w/o} $\mathcal{L}_{\text{spatial}}$} & {\scriptsize{w/o} $\mathcal{L}_{\text{GFL}}$} & {\scriptsize{w/o} prior} & {~Ours~~}\\
            \hline
            Coarse & \textbf{89.3} & 88.9 & 88.5 & 83.4 & 88.8 & 89.1\\
            Fine & 80.8 & 74.1 & 78.5 & 42.3 & 78.3 & \textbf{84.3}\\
            All & 83.2 \scriptsize{$(-2.6\%)$} & 80.3 \scriptsize{$(-6.0\%)$} & 82.0 \scriptsize{$(-4.0\%)$} &
            58.6 \scriptsize{$(-31.3\%)$} & 82.1 \scriptsize{$(-3.9\%)$} & \textbf{85.4}\\
        \end{tabular}
    \end{center}
    \vspace{-5px}
    \caption{
    Ablation study to evaluate the contribution of each component. 
    We remove granularity prior (w/o~prior), GFL loss (w/o~$\mathcal{L}_{\text{GFL}}$), and regularization losses (w/o~$\mathcal{L}_{\text{2D-norm}}$, w/o~$\mathcal{L}_{\text{3D-norm}}$, and w/o~$\mathcal{L}_{\text{spatial}}$) from our complete method to assess each component's impact.
    Average mIoU values on LERF-Mask Dataset reported.
    }
    \label{table:ablation}
\vspace{-15px}
\end{table}

\subsection{Ablation Study}
\label{sec:ablation}
We evaluated the impact of removing granularity prior, GFL loss, and regularization terms from Click-Gaussian.
Without the granularity prior (w/o prior), the model learns two-level features independently, lacking collaborative enhancement.
As Tab.~\ref{table:ablation} shows, our complete model outperformed the one without prior in fine-level segmentation.
This highlights how the intrinsic dependency between coarse and fine levels aids fine-level feature learning.
GFL loss significantly improves fine-level segmentation by effectively addressing inconsistency and ambiguities of SAM-generated fine-level masks.
Removing each regularization loss confirmed their collective importance in enhancing segmentation performance.

\begin{figure*}[t!]
\begin{center}
\centerline{\includegraphics[width=0.99\textwidth]
{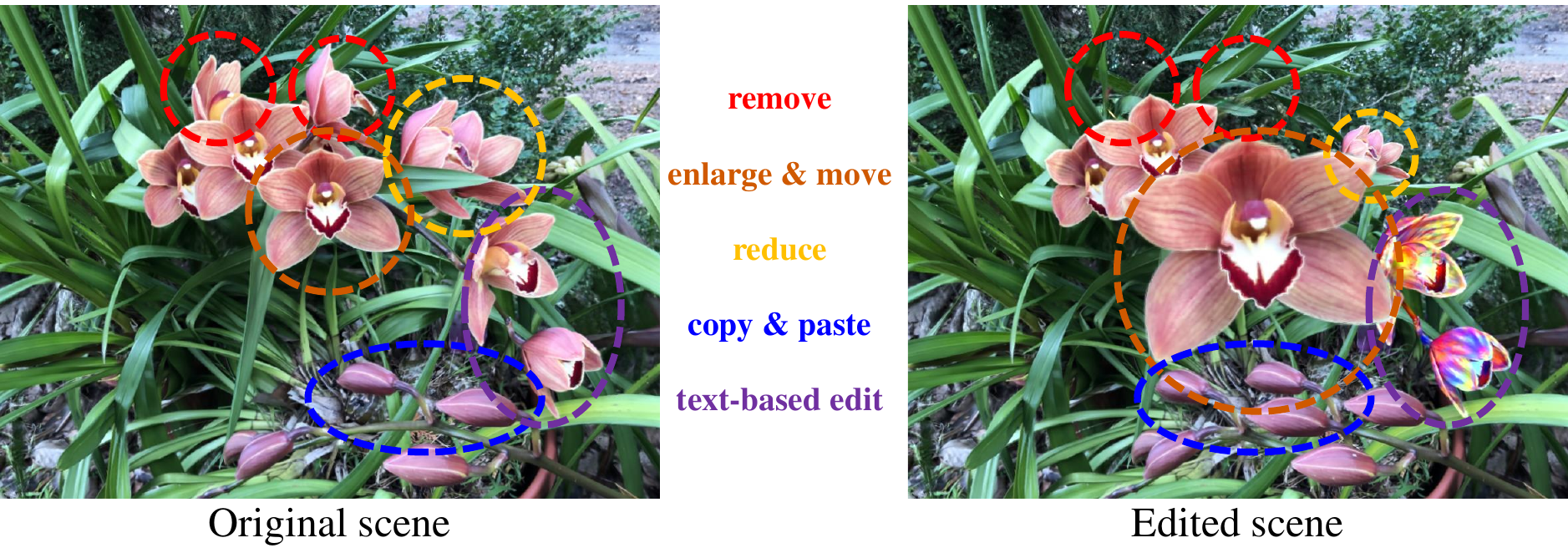}}
\vskip 0.1in
\vspace{-10px}
\caption{
Application of Click-Gaussian. Click-Gaussian enables interactive object selection within about 10ms for various edits. For text-based editing, we utilized CLIP-based editing methods with the source text \textit{`flower'} and target text \textit{`stained glass flower'}.
}
\vspace{-25px}
\label{fig:application}
\end{center}
\end{figure*}
  
\subsection{Versatile Applications}
After training Click-Gaussian's two-level feature fields, various scene manipulation tasks become feasible, including object removal, resizing, repositioning, duplication, and text-based editing.
As Fig.~\ref{fig:application} shows, the two-level global clusters enable rapid object selection within a scene (approximately 10 ms), facilitating interactive local adjustments to selected Gaussians.
Additionally, for text-based editing, we leveraged CLIP-based methods~\cite{radford2021learning, wang2022clip, song2023blending} to get faster results (within about 10~s) than diffusion-based methods~\cite{rombach2022high, brooks2023instructpix2pix, instructnerf2023}.

\section{Conclusion}
We present Click-Gaussian, a swift and precise method enabling interactive fine-grained segmentation of pre-trained 3D Gaussians by lifting 2D segmentation masks into 3D feature fields of two-level granularity.
Noticing from the intrinsic dependency between coarse and fine levels in the real world, we employ a granularity prior for feature division in the representation of feature fields.
To address feature learning hindered by the cross-view inconsistency masks, an inherent issue in lifting 2D masks to 3D, we propose the Global Feature-guided Learning method for more consistent feature field training.
Once Click-Gaussian is trained, users can select desired objects at coarse and fine levels more swiftly than previous methods.
This enhanced capability has the potential to improve efficient and precise 3D environment modification across various applications.

\subsubsection{Limitations.}
Our approach faces limitations due to its reliance on pre-trained 3DGS and the two-level granularity assumption.
Feature learning may be hindered if a single Gaussian represents multiple objects, particularly when they are semantically distinct but chromatically similar.
The two-level granularity assumption, lacking intermediate levels, could limit efficiency for varying granular levels and complex structures, potentially requiring multiple interactions to select desired segmentation regions.


%
%
\bibliographystyle{splncs04}
\bibliography{egbib}

\begin{thebibliography}{10}
\providecommand{\url}[1]{\texttt{#1}}
\providecommand{\urlprefix}{URL }
\providecommand{\doi}[1]{https://doi.org/#1}

\bibitem{kdtree_paper}
Bentley, J.L.: Multidimensional binary search trees used for associative searching. Commun. ACM  \textbf{18}(9),  509–517 (sep 1975). \doi{10.1145/361002.361007}, \url{https://doi.org/10.1145/361002.361007}

\bibitem{brooks2023instructpix2pix}
Brooks, T., Holynski, A., Efros, A.A.: Instructpix2pix: Learning to follow image editing instructions. In: Proceedings of the IEEE/CVF Conference on Computer Vision and Pattern Recognition. pp. 18392--18402 (2023)

\bibitem{videoworldsimulators2024}
Brooks, T., Peebles, B., Holmes, C., DePue, W., Guo, Y., Jing, L., Schnurr, D., Taylor, J., Luhman, T., Luhman, E., Ng, C., Wang, R., Ramesh, A.: Video generation models as world simulators  (2024), \url{https://openai.com/research/video-generation-models-as-world-simulators}

\bibitem{caron2021emerging}
Caron, M., Touvron, H., Misra, I., J{\'e}gou, H., Mairal, J., Bojanowski, P., Joulin, A.: Emerging properties in self-supervised vision transformers. In: Proceedings of the IEEE/CVF international conference on computer vision. pp. 9650--9660 (2021)

\bibitem{cen2024segment3dgaussians}
Cen, J., Fang, J., Yang, C., Xie, L., Zhang, X., Shen, W., Tian, Q.: Segment any 3d gaussians (2024), \url{https://arxiv.org/abs/2312.00860v1}

\bibitem{cen2024segment}
Cen, J., Zhou, Z., Fang, J., Shen, W., Xie, L., Jiang, D., Zhang, X., Tian, Q., et~al.: Segment anything in 3d with nerfs. Advances in Neural Information Processing Systems  \textbf{36} (2024)

\bibitem{chen2023interactive}
Chen, X., Tang, J., Wan, D., Wang, J., Zeng, G.: Interactive segment anything nerf with feature imitation. arXiv preprint arXiv:2305.16233  (2023)

\bibitem{chen2022mobilenerf}
Chen, Z., Funkhouser, T., Hedman, P., Tagliasacchi, A.: Mobilenerf: Exploiting the polygon rasterization pipeline for efficient neural field rendering on mobile architectures. In: The Conference on Computer Vision and Pattern Recognition (CVPR) (2023)

\bibitem{chen2023text}
Chen, Z., Wang, F., Liu, H.: Text-to-3d using gaussian splatting. arXiv preprint arXiv:2309.16585  (2023)

\bibitem{cheng2023tracking}
Cheng, H.K., Oh, S.W., Price, B., Schwing, A., Lee, J.Y.: Tracking anything with decoupled video segmentation. In: Proceedings of the IEEE/CVF International Conference on Computer Vision. pp. 1316--1326 (2023)

\bibitem{Cotton_2024_WACV}
Cotton, R.J., Peyton, C.: Dynamic gaussian splatting from markerless motion capture reconstruct infants movements. In: Proceedings of the IEEE/CVF Winter Conference on Applications of Computer Vision (WACV) Workshops. pp. 60--68 (January 2024)

\bibitem{yu2022plenoxels}
Fridovich-Keil, S., Yu, A., Tancik, M., Chen, Q., Recht, B., Kanazawa, A.: Plenoxels: Radiance fields without neural networks. In: CVPR (2022)

\bibitem{isrfgoel2023}
Goel, R., Sirikonda, D., Saini, S., Narayanan, P.: {Interactive Segmentation of Radiance Fields}. In: {Proceedings of the IEEE/CVF Conference on Computer Vision and Pattern Recognition (CVPR)} (2023)

\bibitem{instructnerf2023}
Haque, A., Tancik, M., Efros, A., Holynski, A., Kanazawa, A.: Instruct-nerf2nerf: Editing 3d scenes with instructions. In: Proceedings of the IEEE/CVF International Conference on Computer Vision (2023)

\bibitem{DearPyGui}
Hoffstadt, J., Cothren, P., Contributors: Dearpygui. \url{https://github.com/hoffstadt/DearPyGui}

\bibitem{jiang2024vr-gs}
Jiang, Y., Yu, C., Xie, T., Li, X., Feng, Y., Wang, H., Li, M., Lau, H., Gao, F., Yang, Y., Jiang, C.: Vr-gs: A physical dynamics-aware interactive gaussian splatting system in virtual reality. arXiv preprint arXiv:2401.16663  (2024)

\bibitem{kerbl3Dgaussians}
Kerbl, B., Kopanas, G., Leimk{\"u}hler, T., Drettakis, G.: 3d gaussian splatting for real-time radiance field rendering. ACM Transactions on Graphics  \textbf{42}(4) (July 2023), \url{https://repo-sam.inria.fr/fungraph/3d-gaussian-splatting/}

\bibitem{kerr2023lerf}
Kerr, J., Kim, C.M., Goldberg, K., Kanazawa, A., Tancik, M.: Lerf: Language embedded radiance fields. In: Proceedings of the IEEE/CVF International Conference on Computer Vision. pp. 19729--19739 (2023)

\bibitem{kim2024garfield}
Kim, C.M., Wu, M., Kerr, J., Goldberg, K., Tancik, M., Kanazawa, A.: Garfield: Group anything with radiance fields (2024)

\bibitem{ICCV2023_SAM}
Kirillov, A., Mintun, E., Ravi, N., Mao, H., Rolland, C., Gustafson, L., Xiao, T., Whitehead, S., Berg, A.C., Lo, W.Y., Dollar, P., Girshick, R.: Segment anything. In: Proceedings of the IEEE/CVF International Conference on Computer Vision (ICCV). pp. 4015--4026 (October 2023)

\bibitem{Tanks-and-Temples}
Knapitsch, A., Park, J., Zhou, Q.Y., Koltun, V.: Tanks and temples: benchmarking large-scale scene reconstruction. ACM Trans. Graph.  \textbf{36}(4) (jul 2017). \doi{10.1145/3072959.3073599}, \url{https://doi.org/10.1145/3072959.3073599}

\bibitem{kobayashi2022distilledfeaturefields}
Kobayashi, S., Matsumoto, E., Sitzmann, V.: Decomposing nerf for editing via feature field distillation. In: Advances in Neural Information Processing Systems. vol.~35 (2022), \url{https://arxiv.org/pdf/2205.15585.pdf}

\bibitem{kopanas2022neural}
Kopanas, G., Leimk{\"u}hler, T., Rainer, G., Jambon, C., Drettakis, G.: Neural point catacaustics for novel-view synthesis of reflections. ACM Transactions on Graphics (TOG)  \textbf{41}(6),  1--15 (2022)

\bibitem{kopanas2021point}
Kopanas, G., Philip, J., Leimk{\"u}hler, T., Drettakis, G.: Point-based neural rendering with per-view optimization. In: Computer Graphics Forum. vol.~40, pp. 29--43. Wiley Online Library (2021)

\bibitem{ling2023align}
Ling, H., Kim, S.W., Torralba, A., Fidler, S., Kreis, K.: Align your gaussians: Text-to-4d with dynamic 3d gaussians and composed diffusion models. arXiv preprint arXiv:2312.13763  (2023)

\bibitem{liu2023zero}
Liu, R., Wu, R., Van~Hoorick, B., Tokmakov, P., Zakharov, S., Vondrick, C.: Zero-1-to-3: Zero-shot one image to 3d object. In: Proceedings of the IEEE/CVF International Conference on Computer Vision. pp. 9298--9309 (2023)

\bibitem{luiten2023dynamic}
Luiten, J., Kopanas, G., Leibe, B., Ramanan, D.: Dynamic 3d gaussians: Tracking by persistent dynamic view synthesis. arXiv preprint arXiv:2308.09713  (2023)

\bibitem{HDBSCAN_paper}
McInnes, L., Healy, J., Astels, S.: hdbscan: Hierarchical density based clustering. Journal of Open Source Software  \textbf{2}(11), ~205 (2017). \doi{10.21105/joss.00205}, \url{https://doi.org/10.21105/joss.00205}

\bibitem{mildenhall2019local}
Mildenhall, B., Srinivasan, P.P., Ortiz-Cayon, R., Kalantari, N.K., Ramamoorthi, R., Ng, R., Kar, A.: Local light field fusion: Practical view synthesis with prescriptive sampling guidelines. ACM Transactions on Graphics (TOG)  \textbf{38}(4),  1--14 (2019)

\bibitem{mildenhall2021nerf}
Mildenhall, B., Srinivasan, P.P., Tancik, M., Barron, J.T., Ramamoorthi, R., Ng, R.: Nerf: Representing scenes as neural radiance fields for view synthesis. Communications of the ACM  \textbf{65}(1),  99--106 (2021)

\bibitem{mirzaei2023spin}
Mirzaei, A., Aumentado-Armstrong, T., Derpanis, K.G., Kelly, J., Brubaker, M.A., Gilitschenski, I., Levinshtein, A.: Spin-nerf: Multiview segmentation and perceptual inpainting with neural radiance fields. In: Proceedings of the IEEE/CVF Conference on Computer Vision and Pattern Recognition. pp. 20669--20679 (2023)

\bibitem{radford2021learning}
Radford, A., Kim, J.W., Hallacy, C., Ramesh, A., Goh, G., Agarwal, S., Sastry, G., Askell, A., Mishkin, P., Clark, J., et~al.: Learning transferable visual models from natural language supervision. In: International conference on machine learning. pp. 8748--8763. PMLR (2021)

\bibitem{ren2023dreamgaussian4d}
Ren, J., Pan, L., Tang, J., Zhang, C., Cao, A., Zeng, G., Liu, Z.: Dreamgaussian4d: Generative 4d gaussian splatting. arXiv preprint arXiv:2312.17142  (2023)

\bibitem{ren-cvpr2022-nvos}
Ren, Z., Agarwala$^\dagger$, A., Russell$^\dagger$, B., Schwing$^\dagger$, A.G., Wang$^\dagger$, O.: Neural volumetric object selection. In: IEEE/CVF Conference on Computer Vision and Pattern Recognition (CVPR) (2022), ($^\dagger$ alphabetic ordering)

\bibitem{rombach2022high}
Rombach, R., Blattmann, A., Lorenz, D., Esser, P., Ommer, B.: High-resolution image synthesis with latent diffusion models. In: Proceedings of the IEEE/CVF conference on computer vision and pattern recognition. pp. 10684--10695 (2022)

\bibitem{yu_and_fridovichkeil2021plenoxels}
{Sara Fridovich-Keil and Alex Yu}, Tancik, M., Chen, Q., Recht, B., Kanazawa, A.: Plenoxels: Radiance fields without neural networks. In: CVPR (2022)

\bibitem{schoenberger2016sfm}
Sch\"{o}nberger, J.L., Frahm, J.M.: Structure-from-motion revisited. In: Conference on Computer Vision and Pattern Recognition (CVPR) (2016)

\bibitem{schoenberger2016mvs}
Sch\"{o}nberger, J.L., Zheng, E., Pollefeys, M., Frahm, J.M.: Pixelwise view selection for unstructured multi-view stereo. In: European Conference on Computer Vision (ECCV) (2016)

\bibitem{make-sense}
Skalski, P.: {Make Sense}. \url{https://github.com/SkalskiP/make-sense/} (2019)

\bibitem{song2023blending}
Song, H., Choi, S., Do, H., Lee, C., Kim, T.: Blending-nerf: Text-driven localized editing in neural radiance fields. In: Proceedings of the IEEE/CVF International Conference on Computer Vision (ICCV). pp. 14383--14393 (October 2023)

\bibitem{segment-anything-nerf}
Tang, J., Chen, X., Wan, D., Wang, J., Zeng, G.: Segment-anything nerf. \url{https://github.com/ashawkey/Segment-Anything-NeRF} (2023)

\bibitem{tang2023dreamgaussian}
Tang, J., Ren, J., Zhou, H., Liu, Z., Zeng, G.: Dreamgaussian: Generative gaussian splatting for efficient 3d content creation. arXiv preprint arXiv:2309.16653  (2023)

\bibitem{tschernezki22neuralyang2023real}
Tschernezki, V., Laina, I., Larlus, D., Vedaldi, A.: {Neural Feature Fusion Fields}: {3D} distillation of self-supervised {2D} image representations. In: Proceedings of the International Conference on {3D} Vision (3DV) (2022)

\bibitem{wang2022clip}
Wang, C., Chai, M., He, M., Chen, D., Liao, J.: Clip-nerf: Text-and-image driven manipulation of neural radiance fields. In: Proceedings of the IEEE/CVF Conference on Computer Vision and Pattern Recognition. pp. 3835--3844 (2022)

\bibitem{VRNeRF}
Xu, L., Agrawal, V., Laney, W., Garcia, T., Bansal, A., Kim, C., Rota~Bulò, S., Porzi, L., Kontschieder, P., Božič, A., Lin, D., Zollhöfer, M., Richardt, C.: {VR-NeRF}: High-fidelity virtualized walkable spaces. In: SIGGRAPH Asia Conference Proceedings (2023). \doi{10.1145/3610548.3618139}, \url{https://vr-nerf.github.io}

\bibitem{yang2023real}
Yang, Z., Yang, H., Pan, Z., Zhu, X., Zhang, L.: Real-time photorealistic dynamic scene representation and rendering with 4d gaussian splatting. arXiv preprint arXiv:2310.10642  (2023)

\bibitem{yang2023deformable}
Yang, Z., Gao, X., Zhou, W., Jiao, S., Zhang, Y., Jin, X.: Deformable 3d gaussians for high-fidelity monocular dynamic scene reconstruction. arXiv preprint arXiv:2309.13101  (2023)

\bibitem{ye2023gaussian}
Ye, M., Danelljan, M., Yu, F., Ke, L.: Gaussian grouping: Segment and edit anything in 3d scenes. arXiv preprint arXiv:2312.00732  (2023)

\bibitem{yen2022nerfsupervision}
Yen-Chen, L., Florence, P., Barron, J.T., Lin, T.Y., Rodriguez, A., Isola, P.: {NeRF-Supervision}: Learning dense object descriptors from neural radiance fields. In: IEEE Conference on Robotics and Automation ({ICRA}) (2022)

\bibitem{yi2023gaussiandreamer}
Yi, T., Fang, J., Wu, G., Xie, L., Zhang, X., Liu, W., Tian, Q., Wang, X.: Gaussiandreamer: Fast generation from text to 3d gaussian splatting with point cloud priors. arXiv preprint arXiv:2310.08529  (2023)

\bibitem{yifan2019differentiable}
Yifan, W., Serena, F., Wu, S., {\"O}ztireli, C., Sorkine-Hornung, O.: Differentiable surface splatting for point-based geometry processing. ACM Transactions on Graphics (TOG)  \textbf{38}(6),  1--14 (2019)

\bibitem{ying2023omniseg3d}
Ying, H., Yin, Y., Zhang, J., Wang, F., Yu, T., Huang, R., Fang, L.: Omniseg3d: Omniversal 3d segmentation via hierarchical contrastive learning (2023)

\bibitem{zhou2023feature}
Zhou, S., Chang, H., Jiang, S., Fan, Z., Zhu, Z., Xu, D., Chari, P., You, S., Wang, Z., Kadambi, A.: Feature 3dgs: Supercharging 3d gaussian splatting to enable distilled feature fields. arXiv preprint arXiv:2312.03203  (2023)

\bibitem{zielonka2023drivable}
Zielonka, W., Bagautdinov, T., Saito, S., Zollh{\"o}fer, M., Thies, J., Romero, J.: Drivable 3d gaussian avatars. arXiv preprint arXiv:2311.08581  (2023)

\end{thebibliography}

\clearpage
\appendix

\begin{center}
\large \textbf{Supplementary Material of \\ Click-Gaussian: Interactive Segmentation \\ to Any 3D Gaussians}
\end{center}

\section{GUI-based Implementation for Click-Gaussian}
\begin{figure*}[h!]
\begin{center}
\centerline{\includegraphics[width=0.85\textwidth]
{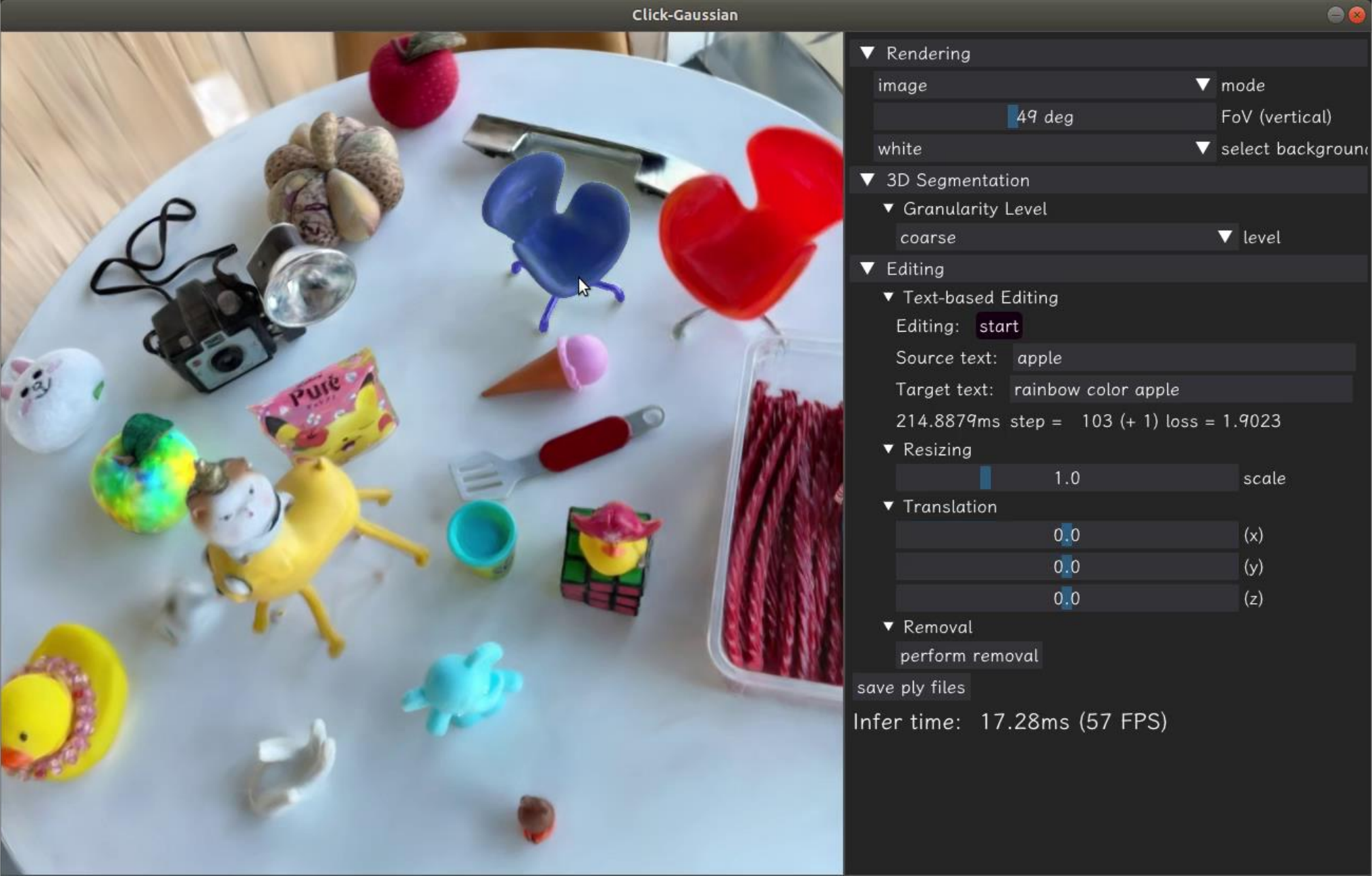}}
\vskip 0.1in
\vspace{-10px}
\caption{
Graphical User Interface (GUI) for Click-Gaussian.
}
\vspace{-33px}
\label{fig:gui}
\end{center}
\end{figure*}

To showcase interactive segmentation and manipulation using Click-Gaussian, we design a Graphical User Interface (GUI) tool based on DearPyGui~\cite{DearPyGui, segment-anything-nerf}, a fast and powerful GUI toolkit for Python. 
As shown in Fig.~\ref{fig:gui}, our GUI is designed to allow users to easily click and segment objects at coarse and fine levels, and provides tools for real-time manipulation tasks such as resizing, translation, removal, and text-based editing for intuitive interaction with the segmented objects.
The supplementary video demonstrates the effectiveness of our method in enabling real-time interactive scene manipulation, showcasing its fast and precise 3D segmentation performance.
We encourage readers to view this video for a comprehensive understanding of the proposed approach's capabilities.

\clearpage

\section{SAM-based Multi-level Mask Generation}
We utilized the official code's \textit{automatic mask generation module} for SAM mask creation, which extracts masks without distinguishing levels, allowing us to get only the highest-confidence segments in an image.
These segments are then assigned to two masks by area: if multiple segments are assigned to a single pixel, the coarse-level mask prioritizes the identity of the larger segment, while the fine-level mask favors the identity of the smaller segments.
This approach enables us to assign a single mask identity per pixel at each level, facilitating stable contrastive learning.

\subsubsection{Comparative Analysis of Multi-level Mask Strategies.}
\begin{figure}[h!]
\begin{center}
\centerline{\includegraphics[width=0.99\columnwidth]{
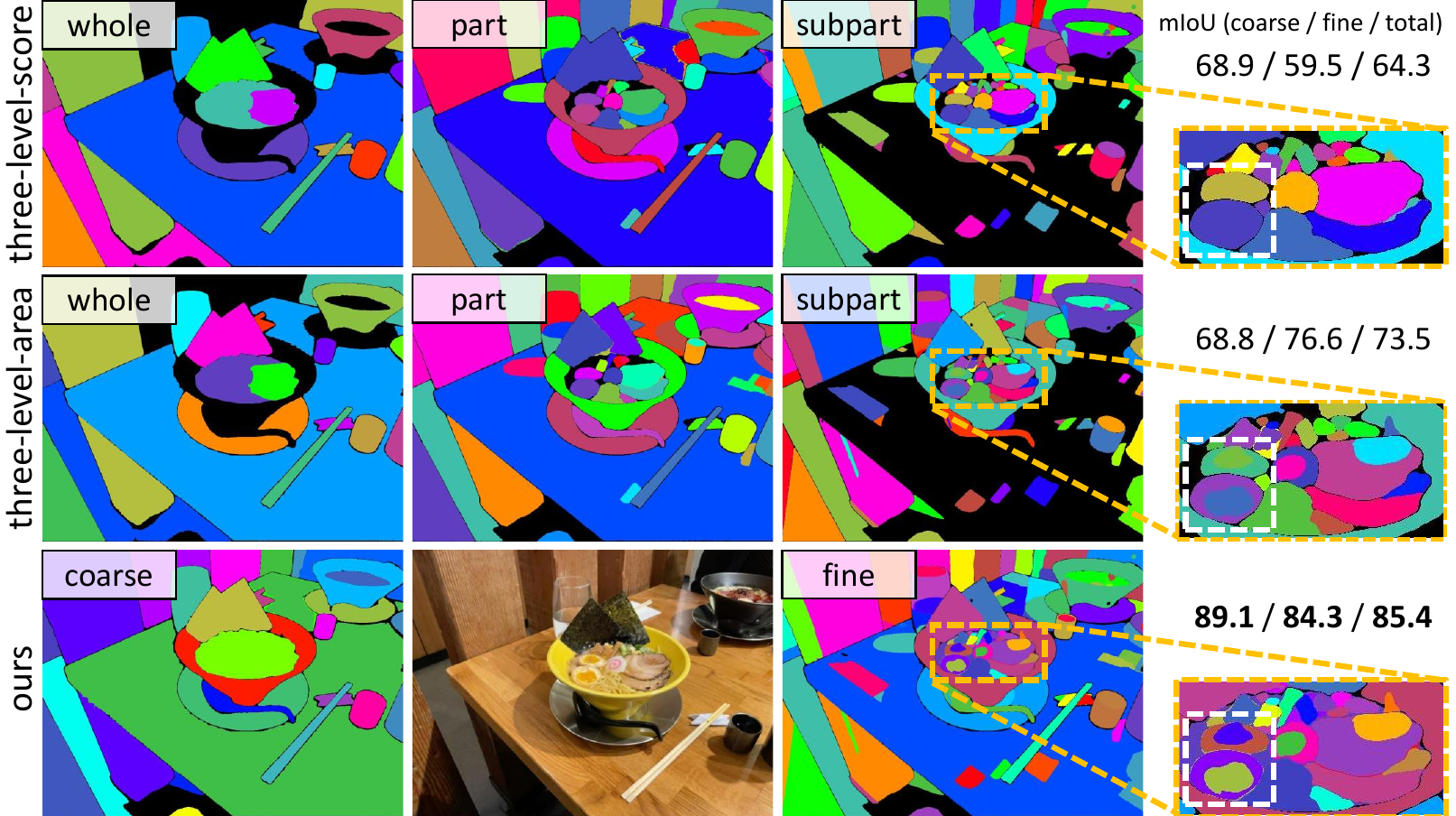}}
\caption{
Performance comparison of different mask generation strategies.
Black areas indicate pixels with unassigned identities.
}
\label{fig:mask}
\end{center}
\end{figure}

Our method can adopt SAM's three-level masks (whole, part, and subpart) in two ways: three-level-score  and three-level-area.
Each approach prioritizes the highest score segment and smallest segment, respectively, for each level.
In these cases, we split $\mathbf{f}_i \in \mathbb{R}^{24}$ into three levels of granularity.
As shown in Fig.~\ref{fig:mask}, the three-level-area outperforms the three-level-score in fine-level mIoU due to finer-grained mask supervision (\eg, egg white and yolk), demonstrating the efficacy of the area-based prioritization.
Additionally, our method using two-level masks surpasses the three-level-area thanks to the mask completeness and training efficiency:
It has fewer unassigned identities than the three-level-area and learns feature fields more efficiently with the same feature dimension of 24.
For these advantages, we adopt the two-level granularity assumption.

\section{Annotations for Evaluating Fine-grained Segmentation}
\begin{figure*}[h!]
\begin{center}
\centerline{\includegraphics[width=0.95\textwidth]
{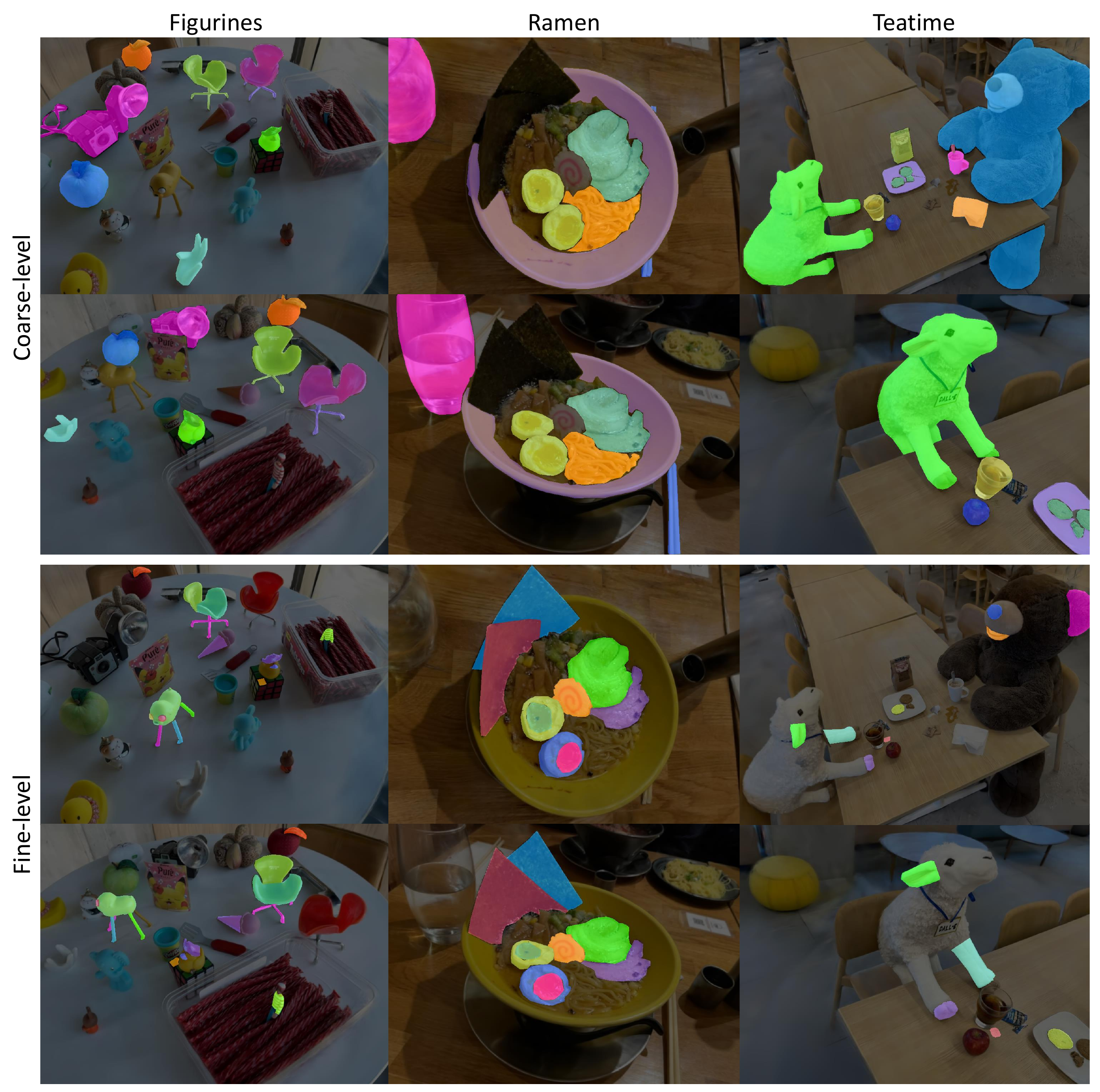}}
\vskip 0.1in
\vspace{-10px}
\caption{
Annotations for evaluating fine-grained segmentation.
The first two rows of each scene show the ground truth annotations for evaluating coarse-level segmentation with two sampled test views.
On the other hand, each scene's last two rows show the ground truth annotations for evaluating fine-level segmentation.
}
\vspace{-33px}
\label{fig:lerf_mask_data_annotation}
\end{center}
\end{figure*}

We evaluate our approach's segmentation performance using the LERF-Mask dataset~\cite{ye2023gaussian}, a public real-world dataset for 3D segmentation tasks.
This dataset comprises three scenes (Figurines, Ramen, and Teatime)~\cite{kerr2023lerf} with manually annotated ground truth masks for semantically large objects, as shown in the first two rows of Fig.~\ref{fig:lerf_mask_data_annotation}.
To assess fine-grained segmentation performance, we additionally annotated masks for smaller objects within each scene using Make-Sense~\cite{make-sense}, a free online image labeling tool, as shown in the last two rows of Fig.~\ref{fig:lerf_mask_data_annotation}.
This additional annotation is necessary due to the lack of datasets suitable for fine-level comparison.
Note that the annotation process was conducted independently from our segmentation experiments.

\section{Additional Experiments and Results}
\subsection{3D Editing in AI-generated Videos}
\begin{figure*}[h!]
\begin{center}
\centerline{\includegraphics[width=0.99\textwidth]
{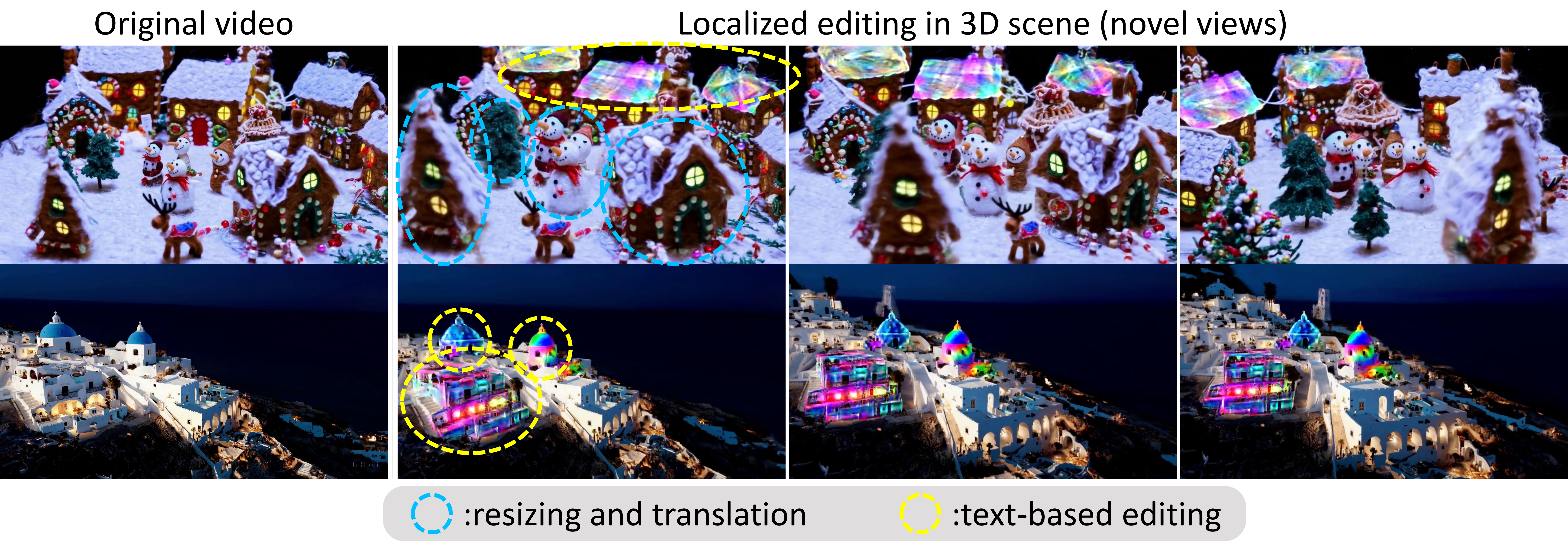}}
\vskip 0.1in
\vspace{-10px}
\caption{
Versatile applications of Click-Gaussian on synthetic videos generated by Sora.
After pre-training 3DGS on Sora-generated videos, users can flexibly modify the reconstructed 3D scenes in real-time, including resizing, translation (\textcolor{skyblue}{sky blue circle}), and text-based editing (\textcolor{yellow}{yellow circle}).
In the Snow-village scene (first row), we manipulated the scene by enlarging and translating three snowmen, two houses, and a tree, while stylizing other house roofs to crystal.
In the Santorini scene (second row), we applied text-based editing to buildings, transforming them into cyberpunk neon, crystal, and rainbow styles from the bottom left, respectively.
}
\vspace{-33px}
\label{fig:sora_edit}
\end{center}
\end{figure*}

OpenAI recently announced Sora~\cite{videoworldsimulators2024}, a groundbreaking text-to-video generation model, showing a promising path towards building general-purpose world simulators. 
These simulators can be further improved by enabling interactive modification of generated realistic environments through accurate and fast 3D segmentation methods like Click-Gaussian, enhancing their functionality and user interaction capabilities.
To demonstrate Click-Gaussian's versatility in scene segmentation and manipulation on these generated scenes, we applied our method to videos (Santorini\textsuperscript{1}\footnote{\textsuperscript{1}\url{https://cdn.openai.com/sora/videos/santorini.mp4}} and Snow-village\textsuperscript{2})\footnote{\textsuperscript{2}\url{https://cdn.openai.com/tmp/s/interp/b2.mp4}. This video has no official name, so we refer to it as Snow-village.} generated by Sora. 
As shown in Fig.~\ref{fig:sora_edit}, after pre-training 3DGS on each generated video using COLMAP~\cite{schoenberger2016sfm, schoenberger2016mvs}, users can flexibly make desired modifications, resulting in more creative and diverse 3D environments with Click-Gaussian.

\newpage

\subsection{Open-vocabulary 3D Object Localization}
\begin{figure*}[h!]
\begin{center}
\centerline{\includegraphics[width=0.99\textwidth]
{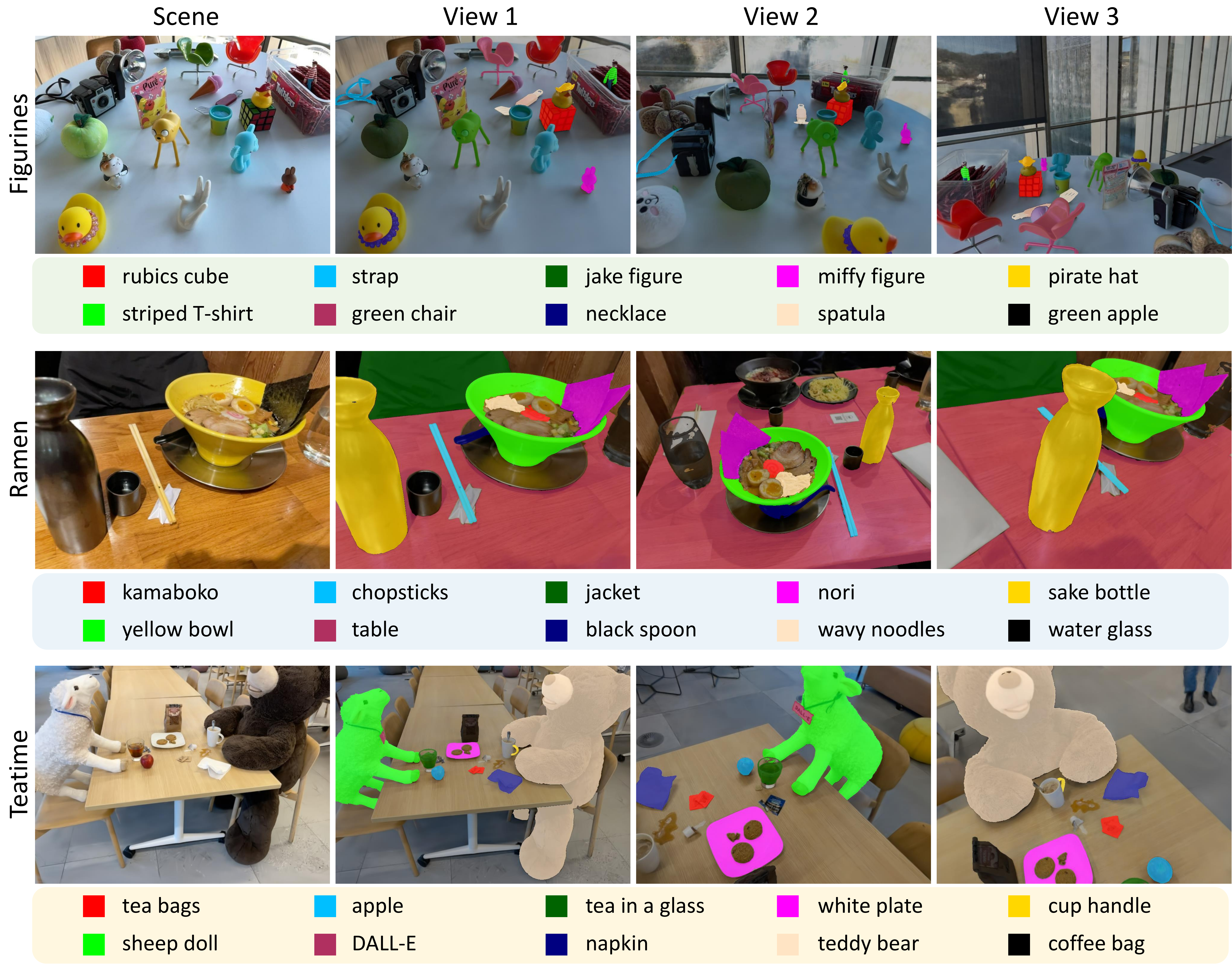}}
\vskip 0.1in
\vspace{-10px}
\caption{
Open-vocabulary 3D object localization results on the LERF-Mask Dataset. 
Segmentation results are color-overlaid for visualization in three different scenes.
}
\vspace{-33px}
\label{fig:text_algin}
\end{center}
\end{figure*}

\begin{figure*}[t!]
\begin{center}
\centerline{\includegraphics[width=0.99\textwidth]
{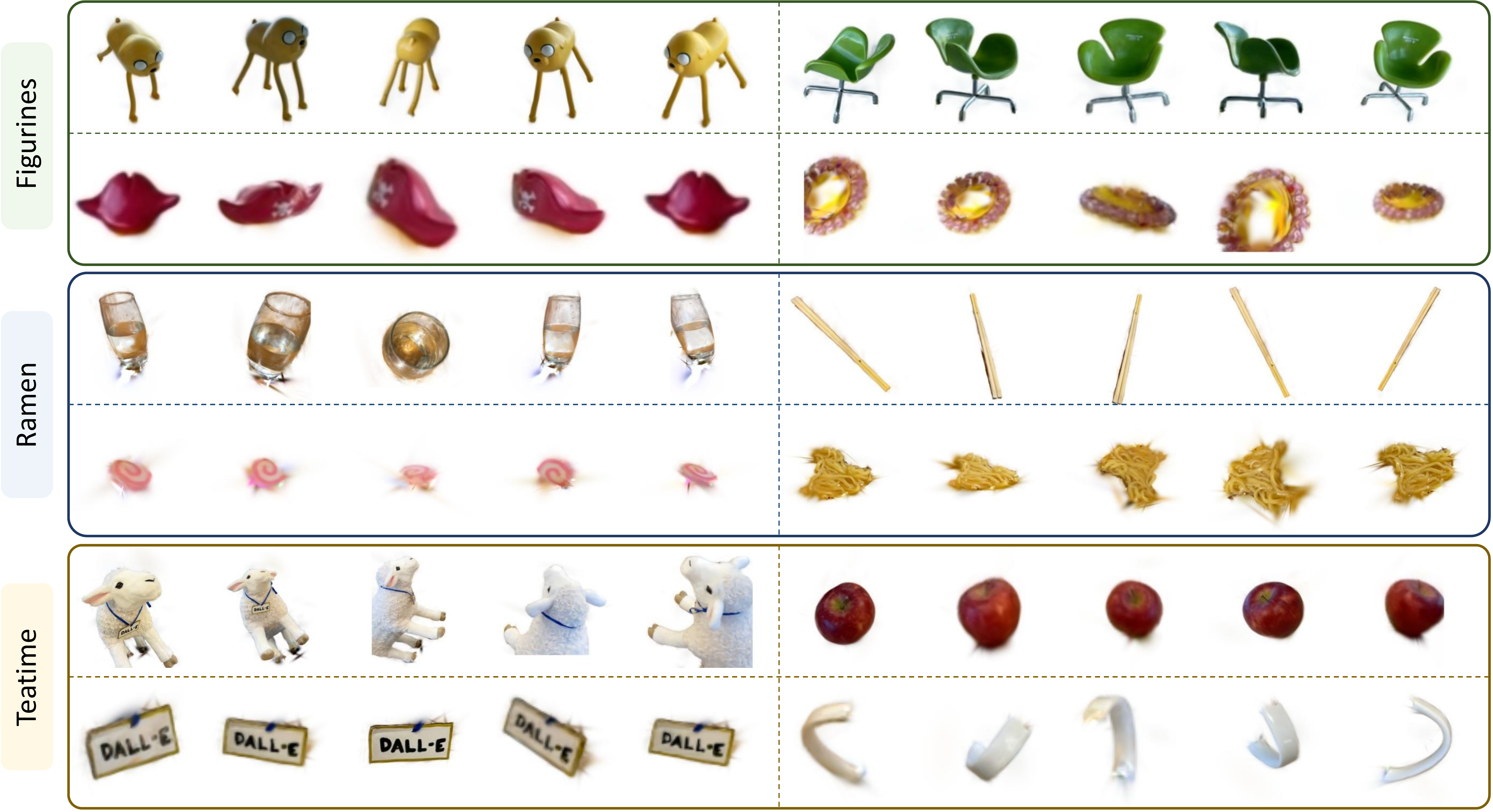}}
\vskip 0.1in
\vspace{-10px}
\caption{
Examples of rendered images using the Gaussians corresponding to each global cluster.
Five representative images are shown per cluster for simplicity.
These images are used to obtain CLIP embeddings for each cluster via the CLIP image encoder.
}
\label{fig:extraction_rendering}
\end{center}
\end{figure*}

Once trained, our method can perform open-vocabulary 3D object localization as shown in Fig.~\ref{fig:text_algin}, using the obtained global feature candidates, which we call global clusters.
Specifically, for all two-level global clusters, we render only the Gaussians corresponding to each cluster in multiple views (10 randomly sampled views) as shown in Fig.~\ref{fig:extraction_rendering}.
We then input these rendered images into the CLIP image encoder~\cite{radford2021learning} to obtain the CLIP embeddings of each cluster.
Thanks to the real-time rendering speed of 3DGS, this process of obtaining CLIP embeddings for all global clusters completes in 20--40 seconds, depending on the number of global clusters in the scene.
Note that this process only needs to be performed once before any text query.
Subsequently, given text queries, open-vocabulary 3D object localization is performed by returning the global cluster with the highest cosine similarity between the obtained image embeddings of all global clusters and the text query embedding.
Fig.~\ref{fig:text_algin} qualitatively demonstrates that our approach precisely localizes 3D objects for given text queries using globally obtained clusters.

\newpage

\subsection{Additional Results for 3D Segmentation}
\subsubsection{Experiments on LeRF Dataset.}
\begin{figure*}[h!]
\begin{center}
\centerline{\includegraphics[width=1.0\textwidth]
{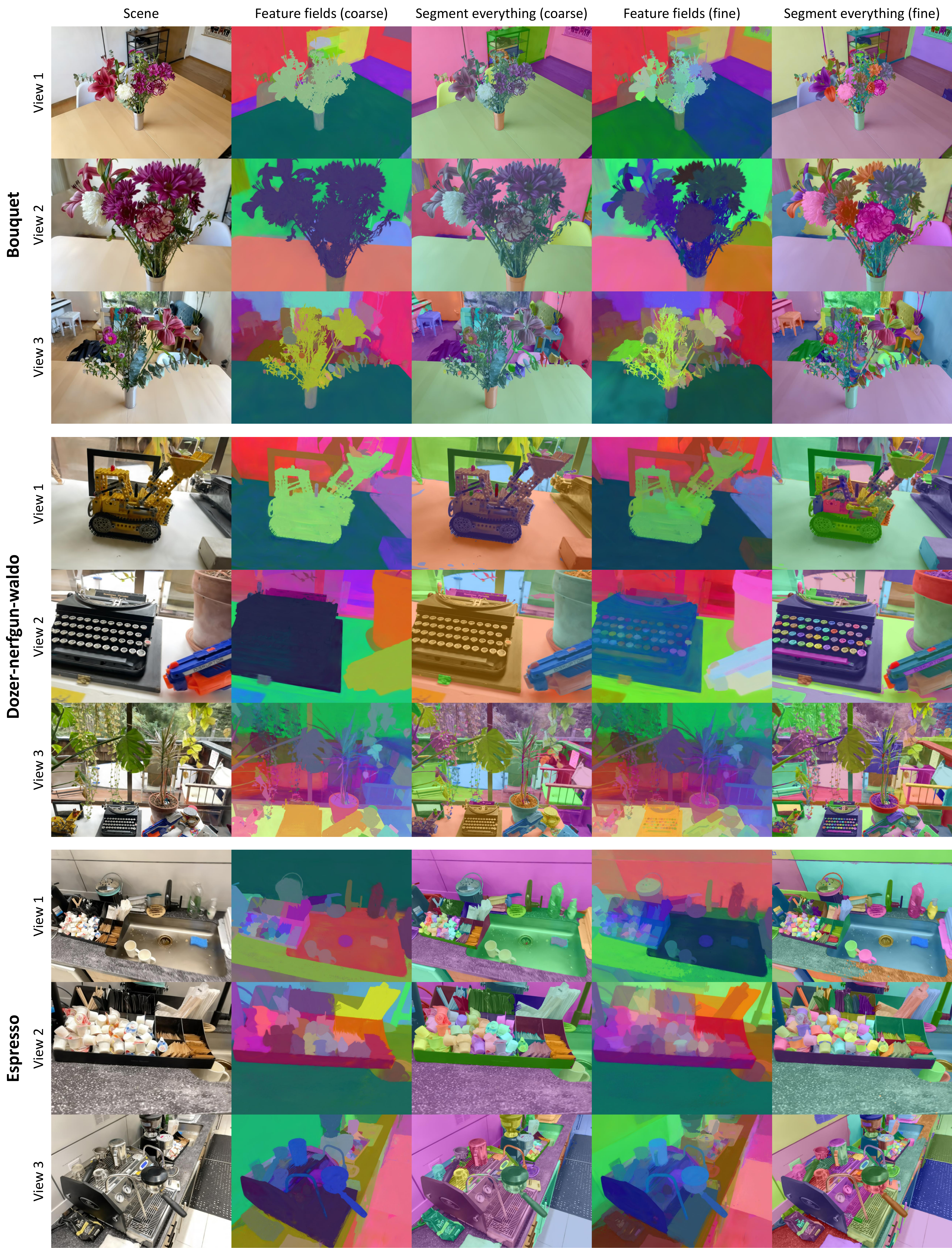}}
\vskip 0.1in
\vspace{-10px}
\caption{
Segmentation of everything results on the LeRF Dataset.
We present automatic segmentation results (third and fifth columns) along with PCA visualizations of rendered feature maps (second and fourth columns) at two granularity levels for Bouquet, Dozer-nerfgun-waldo, and Espresso scenes (first column) from the LeRF Dataset.
Objects classified with the same ID in the segmentation results share the same overlaid color across the three given views, as each global cluster ID remains consistent throughout a scene.
}
\vspace{-33px}
\label{fig:segment_everything_lerf_result1}
\end{center}    
\end{figure*}

\begin{figure*}[h!]
\begin{center}
\centerline{\includegraphics[width=0.99\textwidth]
{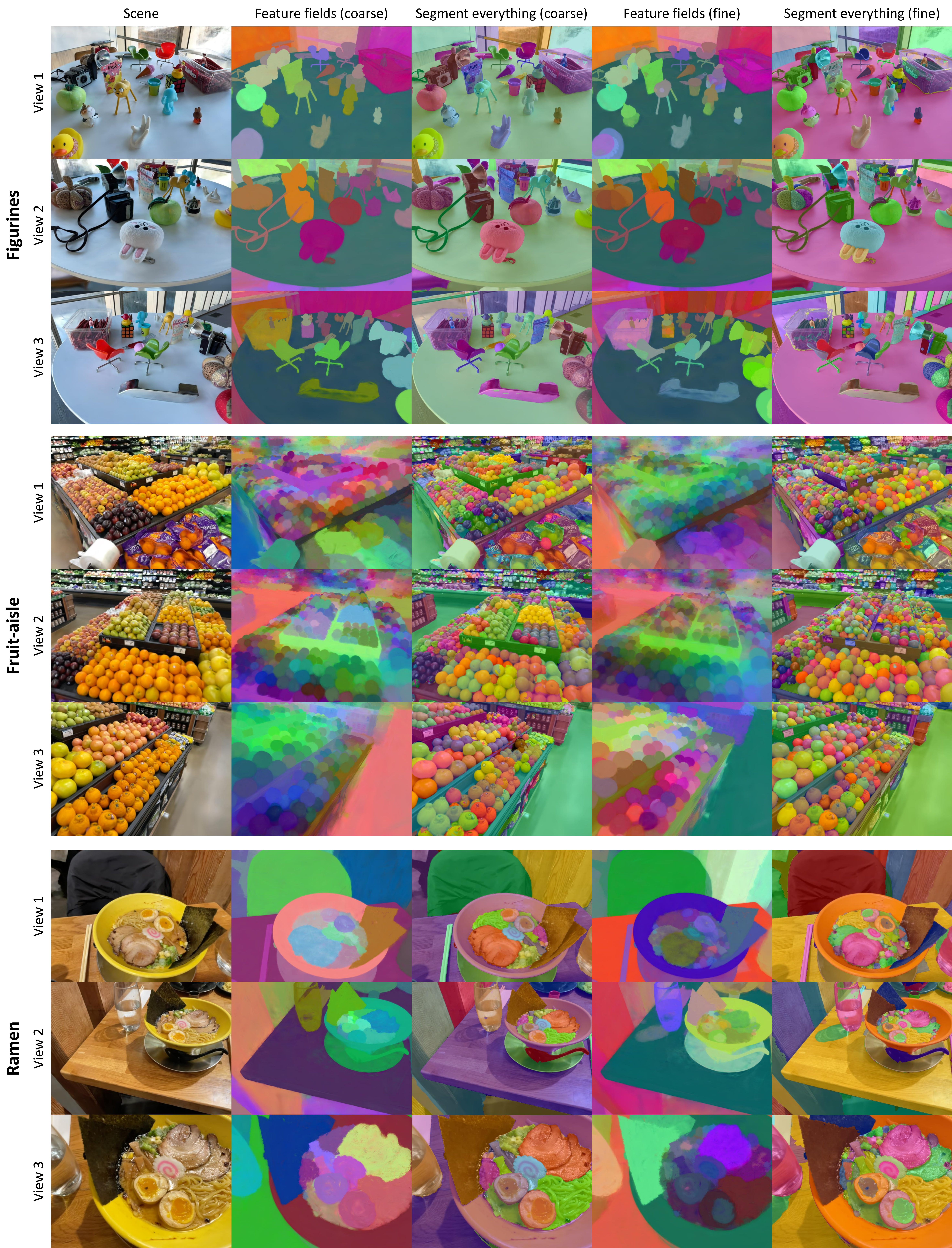}}
\vskip 0.1in
\vspace{-10px}
\caption{
Segmentation of everything results on the LeRF Dataset.
We present automatic segmentation results (third and fifth columns) along with PCA visualizations of rendered feature maps (second and fourth columns) at two granularity levels for Figurines, Fruit-aisle, and Ramen scenes (first column) from the LeRF Dataset.
Objects classified with the same ID in the segmentation results share the same overlaid color across the three given views, as each global cluster ID remains consistent throughout a scene.
}
\vspace{-33px}
\label{fig:segment_everything_lerf_result2}
\end{center}    
\end{figure*}

\begin{figure*}[h!]
\begin{center}
\centerline{\includegraphics[width=1.0\textwidth]
{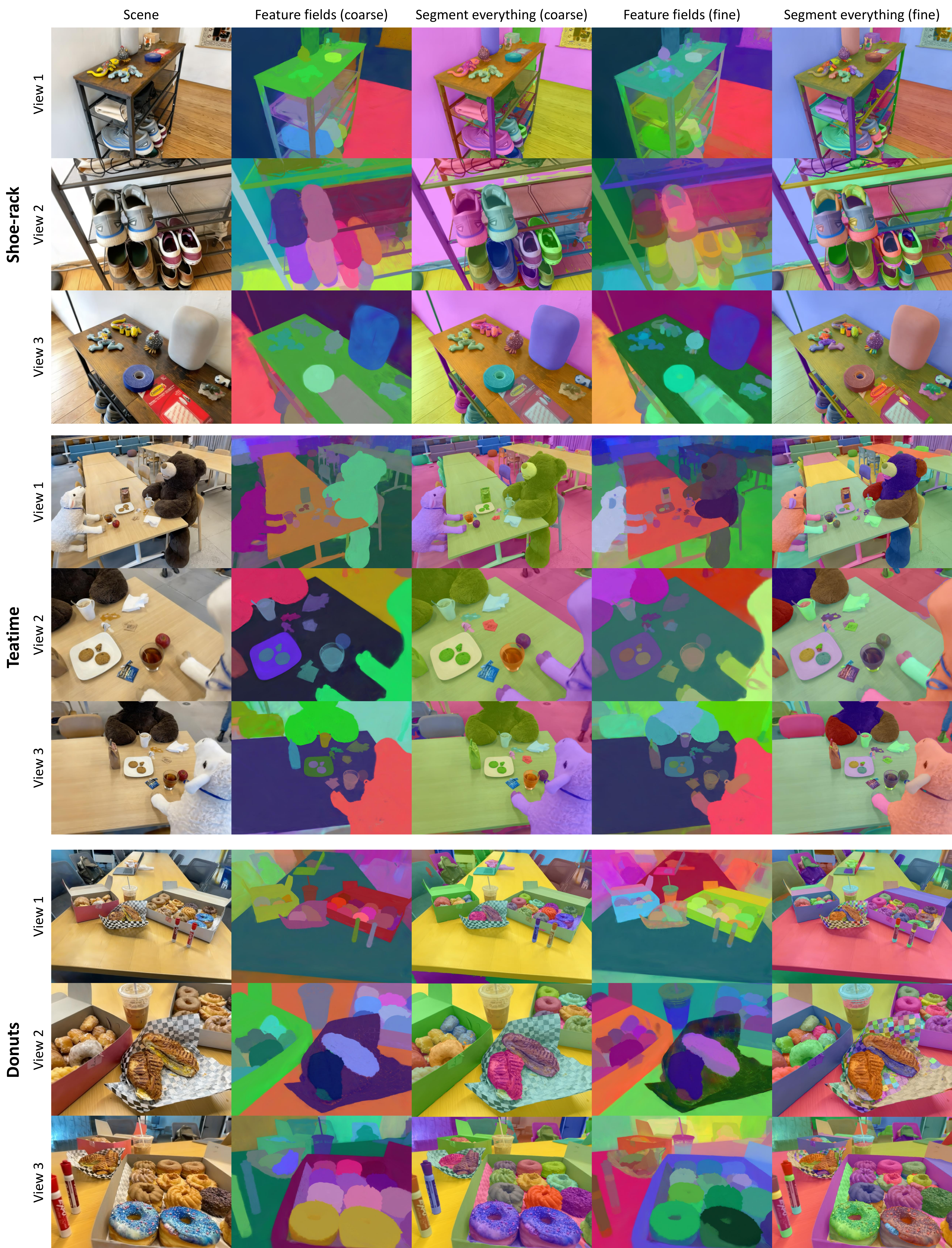}}
\vskip 0.1in
\vspace{-10px}
\caption{
Segmentation of everything results on the LeRF Dataset.
We present automatic segmentation results (third and fifth columns) along with PCA visualizations of rendered feature maps (second and fourth columns) at two granularity levels for Shoe-rack, Teatime, and Donuts scenes (first column) from the LeRF Dataset.
Objects classified with the same ID in the segmentation results share the same overlaid color across the three given views, as each global cluster ID remains consistent throughout a scene.
}
\vspace{-33px}
\label{fig:segment_everything_lerf_result3}
\end{center}    
\end{figure*}

In addition to user-guided segmentation, our approach can also automatically segment everything by calculating the cosine similarity between the rendered 2D feature map and global clusters' features, assigning a global cluster ID with the maximum similarity value to each pixel.
By performing this process for each of the two granularity levels, we obtain automatic segmentation results at both coarse and fine levels.
Figs.~\ref{fig:segment_everything_lerf_result1}, ~\ref{fig:segment_everything_lerf_result2}, and \ref{fig:segment_everything_lerf_result3} show the results of automatic segmentation for several complicated real-world scenes from the LeRF dataset\cite{kerr2023lerf}, along with PCA visualizations of rendered feature maps at two levels.
These results qualitatively demonstrate that Click-Gaussian achieves high-fidelity, fine-grained segmentation of everything in complex real-world scenes.

\subsubsection{Experiments on SPIn-NeRF Dataset.}
\begin{figure*}[h!]
\begin{center}
\centerline{\includegraphics[width=0.95\textwidth]
{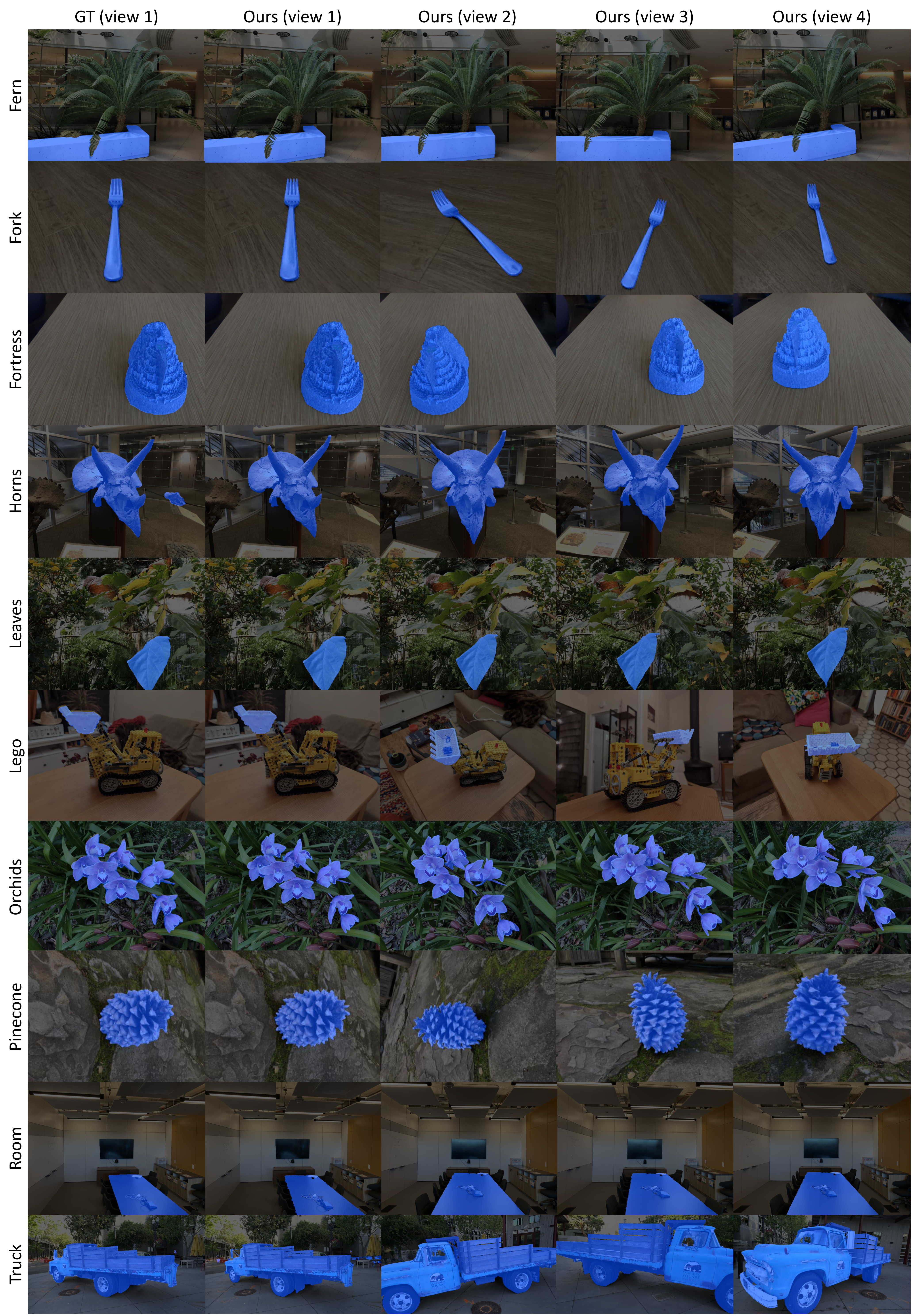}}
\vskip 0.1in
\vspace{-10px}
\caption{
3D segmentation results on the SPIn-NeRF Dataset.
We use the label propagation method based on the ground truth mask of a reference view (first column) to identify the cluster IDs belonging to the target object.
These IDs are then used to generate 2D masks for test views (subsequent columns).
}
\vspace{-33px}
\label{fig:ours_spinnerf_seg3d}
\end{center}    
\end{figure*}

We further showcase the 3D multi-view segmentation results on the SPIn-NeRF Dataset using the label propagation method, as illustrated in Fig.~\ref{fig:ours_spinnerf_seg3d}.
These results offer additional examples demonstrating the effectiveness of Click-Gaussian across various real-world scenes.


\end{document}